\documentclass[10pt,twocolumn,letterpaper]{article}

\usepackage[pagenumbers]{cvpr} %

\usepackage{colortbl}
\usepackage{multirow}
\usepackage[ruled]{algorithm}
\usepackage{algpseudocode}
\usepackage{cuted} %

\usepackage{currfile} %

\usepackage{caption} %

\usepackage{placeins}
\renewcommand{\dbltopfraction}{0.85}
\renewcommand{\textfraction}{0.1}
\renewcommand{\topfraction}{0.85}

\usepackage{fontawesome5}

\definecolor{cvprblue}{rgb}{0.21,0.49,0.74}
\usepackage[pagebackref,breaklinks,colorlinks,allcolors=cvprblue]{hyperref}

\newcommand{\ours}{\texttt{Dooop}}
\newcommand{\vT}{v^{\mathrm{T}}}

\def\confName{CVPR}
\def\confYear{2027}

\title{Distill to Loop: Turning Pretrained Diffusion Transformers \\ into Looped Models with Adjustable Depth}

\author{Chenru Wang \qquad Shengwei An$^{\dagger}$\\
Virginia Tech\\
{\tt\small winterccccc@gmail.com \qquad swan@vt.edu}
}

\begin{document}
\twocolumn[{%
\renewcommand\twocolumn[1][]{#1}%
\maketitle
\vspace{-1.6em}
\begin{center}
{\small
\href{https://github.com/WinterCCC/Distill-to-Loop}{\faGithub\ Code}\qquad
\href{https://huggingface.co/ShimaRinC/Distill-to-Loop}{\raisebox{-0.15em}{\includegraphics[height=1em]{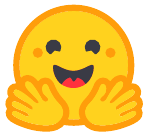}}\ Models}\qquad
\href{https://winterccc.github.io/dooop}{\faHome\ Project page}
}
\end{center}
\vspace{-0.6em}
\begin{center}
\includegraphics[width=\textwidth]{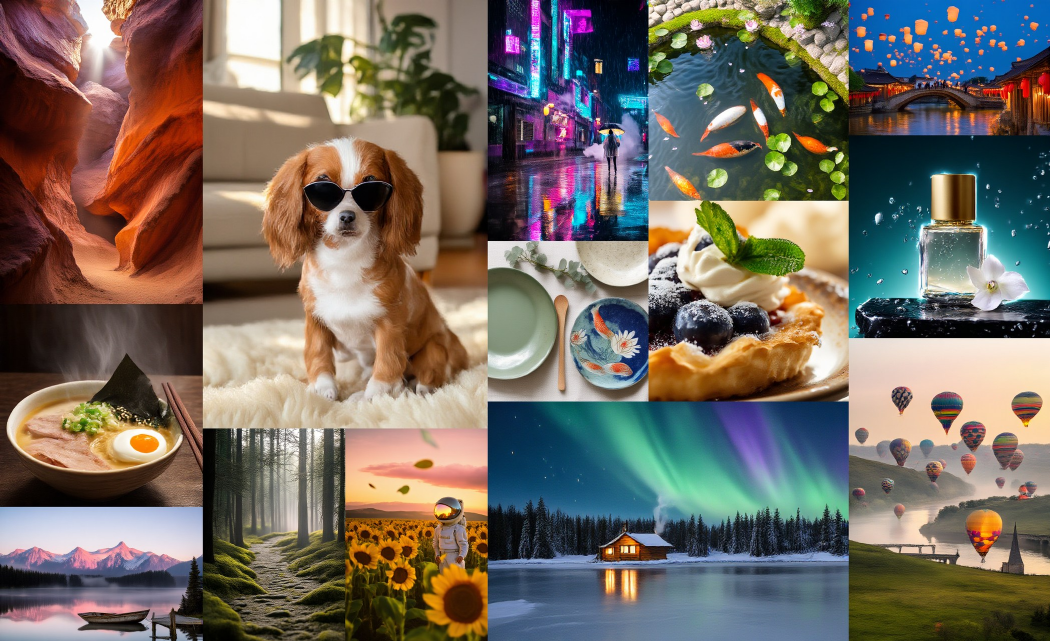}
\captionof{figure}{Samples of the \ours~student distilled from Z-Image.}
\label{fig:teaser_zimage}
\end{center}
\vspace{0.8em}
}]
{\let\thefootnote\relax\footnotetext{$^{\dagger}$Corresponding author.}}
\begin{abstract}
Looped transformers reuse one block of shared weights along the depth of the network. Recent looped diffusion transformers are trained from scratch, which repeats the full pretraining cost although strong pretrained non-looped diffusion transformers exist. We propose Distill to Loop~(\ours), which distills a pretrained non-looped diffusion transformer into a looped student whose blocks are initialized from the teacher's. We cast distillation as a fixed-point iteration toward the teacher, in which the output after every loop is trained toward the teacher, and each loop is one step of an ordinary differential equation in depth. As a result, one model runs any number of loops at test time without retraining, and the higher-order ODE solvers of diffusion model sampling can be applied along the depth dimension for better results. Meanwhile, because a looped model produces an output after every loop, we also construct loop guidance between the outputs of different loop counts, which improves the generated samples at little extra compute. On SD3-medium, a looped student with about 37\% of the teacher's parameters matches the teacher on average over GenEval, ImageReward, PickScore and HPSv2.1 with two to four loops, and with two loops it uses about 58\% of the teacher's compute. And on a four-step SD3-medium, loop guidance raises the average of the four metrics from 88.2 to 99.2.
\end{abstract}

\section{Introduction}
\label{sec:intro}

Looped transformers, which apply a block of shared weights repeatedly along the depth of the network, have recently gained traction in language modeling~\cite{dehghani2019universal,geiping2025scaling,zhu2025scaling}. A looped model can spend more computation at inference time by running more iterations. The accuracy of recurrent-depth language models improves as the number of iterations grows at test time~\cite{geiping2025scaling}, and the looped model Ouro~\cite{zhu2025scaling} approaches the quality of larger dense models. Sharing weights across iterations reduces the parameter count and raises throughput~\cite{bae2025relaxed,bae2025mixture}, and pretrained non-looped language models can be converted into looped ones~\cite{bae2025relaxed,mcleish2025teaching,park2026loopus}.

Diffusion models are among the generative models most limited by inference time, and video diffusion models most of all, because generating one video runs a large backbone, such as the 14B-parameter Wan~\cite{wan2025wan}, for many denoising steps. Step distillation reduces the number of steps, for example through progressive distillation~\cite{salimans2022progressive}, consistency models~\cite{song2023consistency}, distribution matching distillation~\cite{yin2024one,yin2024improved}, and its video extension CausVid~\cite{yin2025slow}. Each remaining step still runs the full backbone, so inference stays long. Looped transformers have recently appeared in visual generation as well. Looped-DiT~\cite{chng2026looped}, ELT~\cite{goyal2026elastic}, and LiFT~\cite{derakhshani2026lift} show that a small set of shared blocks can replace a deep stack of distinct blocks. In these models, running fewer loops gives a faster sample.

But all of these looped diffusion models are trained from scratch on the full dataset. Obtaining a looped model repeats the full pretraining cost, although strong pretrained non-looped diffusion transformers already exist. This leads us to two questions. The first question asks whether a looped model can be obtained faster by distilling an existing pretrained model. And the second question concerns how many loops a looped model needs. A looped model is a fixed-point iteration, so a good one should reach its fixed point within few loops. Some existing looped diffusion transformers~\cite{goyal2026elastic,derakhshani2026lift} reach the quality of a non-looped model only at a large depth, at their full loop count or with more loops than they were trained with. Other models~\cite{chng2026looped} supervise every loop with the final objective, so their early loops converge sooner, but their quality drops once they run more loops than they were trained with. Methods that adapt depth at test time, such as choosing sampling steps per prompt~\cite{zhang2025adadiff}, skipping calls or layers~\cite{ye2024training,binici2026efficient,tang2024adadiff}, routing to pruned experts~\cite{ganjdanesh2025not}, or exiting a loop early~\cite{chng2026looped}, only remove computation. We call this question \emph{adjustable depth}. It asks whether a looped model obtained by distillation can converge within few loops, so that a small loop count already matches the teacher, while it keeps the ability to extend its depth, with the loop count and the depth solver set at inference without retraining.

Building on these two questions, we propose \underline{D}istill t\underline{o} L\underline{oop}~(\ours). \ours~takes a pretrained non-looped diffusion transformer, keeps a few leading and trailing blocks, and replaces the middle stack by a small set of shared blocks that are looped. Every block of the student is initialized from a teacher block, so training starts from the teacher's weights. We cast distillation as a fixed-point iteration in which the student moves toward the teacher loop by loop, so during training the teacher velocity is the fixed point. The target of the output after every loop is the result of one iteration from the previous loop's output toward the teacher. Every loop is therefore pulled toward the teacher, each loop is tied to the loop before it, and the last loop reaches the teacher. This allows a small loop count to already land close to the teacher. In the forward pass, we model the student's loops as an ordinary differential equation in depth over the depth interval from 0 to 1. One step of this equation is a relaxed fixed-point step whose step size is set by a continuous depth coordinate. Every loop count therefore covers the same depth interval, so one model runs any number of loops at inference without retraining. At the same time, because the loops are steps of this equation, we can also apply here the higher-order solvers of diffusion model sampling~\cite{karras2022elucidating,lu2022dpm,lu2025dpm} to obtain a more accurate iteration. As for the training states, \ours~trains not only on noisy data states but also on on-policy states, which a sampler that combines the teacher and the current student visits, so that training covers the states the student meets when it samples. For sampling, we design loop guidance based on the looped structure, which uses the output of an earlier loop in the same forward pass as the weak model and improves the samples at little extra compute.

We evaluate \ours~on class-conditional ImageNet~\cite{deng2009imagenet} generation with SiT-XL/2 as the teacher and mainly on text-to-image generation with SD3-medium.
On SD3-medium, our looped student has 0.756B parameters, about 37\% of the teacher's parameters, and at four loops it costs the same per-image compute as the teacher. With two loops it uses only about 58\% of the teacher's compute and already exceeds the teacher over GenEval, ImageReward, PickScore and HPSv2.1, with an average of 100.7 against the teacher's 100. We also continue training the distilled student on real data. With two loops, this student exceeds the teacher over the four metrics with an average of 105.8. In particular, its GenEval of 0.707 is well above the teacher's 0.648. On a four-step student, loop guidance raises the average over the four metrics relative to the four-step teacher from 88.2 to 99.2, while the compute per image grows only from 7.21 to 7.79 TFLOPs.

Our contributions are summarized as follows:
\begin{itemize}
\item We propose \ours, which distills a pretrained non-looped diffusion transformer into a looped one by casting distillation as a fixed-point iteration toward the teacher, in which the target of the output after every loop is the result of one iteration from the previous loop's output toward the teacher, so that a small loop count already lands close to the teacher.
\item We model the loops of one forward pass as an ordinary differential equation in depth, so that one model runs any number of loops at inference without retraining and can use the higher-order solvers of diffusion sampling, and we design loop guidance, which builds guidance between the current output and the output of an earlier loop in the same forward pass, used as the weak model.
\item On SD3-medium, the looped student with about 37\% of the teacher's parameters exceeds the teacher on average over four metrics with two loops at about 58\% of its compute, continued training on real data raises this average to 105.8, and loop guidance raises the average of a four-step student from 88.2 to 99.2 at little extra compute. Code will be released with the formal version.
\end{itemize}

\section{Related Work}
\label{sec:related}

\noindent\textbf{Looped Transformers.} The Universal Transformer~\cite{dehghani2019universal} applies one shared block repeatedly along depth, and recent language models such as Huginn~\cite{geiping2025scaling} and Ouro~\cite{zhu2025scaling} train this recurrence from scratch. Relaxed Recursive Transformers~\cite{bae2025relaxed}, retrofitted recurrence~\cite{mcleish2025teaching}, LoopUS~\cite{park2026loopus} and ALoDLM~\cite{fang2026adaptively} convert pretrained non-looped language models into looped ones, and GET~\cite{geng2023one} distills a diffusion model into a one-step equilibrium transformer. In diffusion transformers, Looped-DiT~\cite{chng2026looped}, ELT~\cite{goyal2026elastic} and LiFT~\cite{derakhshani2026lift} train looped models from scratch. ELT and LiFT can extend their depth, but they reach the quality of a non-looped baseline only at the same or more compute. Looped-DiT reaches that quality with less compute, but its quality drops once it runs more loops than it was trained with. \ours~distills a pretrained diffusion transformer instead. While it needs less training, it reaches the teacher's quality with less compute and can still extend its depth.

\noindent\textbf{Guidance for Diffusion Models.} Classifier-free guidance~\cite{ho2022classifier} extrapolates from the unconditional prediction toward the conditional one, which needs a second network evaluation without the condition at every step. Autoguidance~\cite{karras2024guiding} extrapolates away from a weak model, a smaller or less-trained version of the same model, which must be trained and stored separately and run as an extra network evaluation. S$^2$-Guidance~\cite{chen2026stochastic} builds the weak model by dropping blocks of the model itself, which removes the separate training but still runs the sub-network as an extra forward pass. Loop guidance in \ours~takes the output of an earlier loop in the same forward pass as the weak model and builds guidance between it and the current output, so it needs no separately trained weak model and no extra forward pass, only one extra evaluation of the output module.

\section{Preliminaries}
\label{sec:prelim}
\begin{figure*}[t]
\centering
\includegraphics[width=\textwidth]{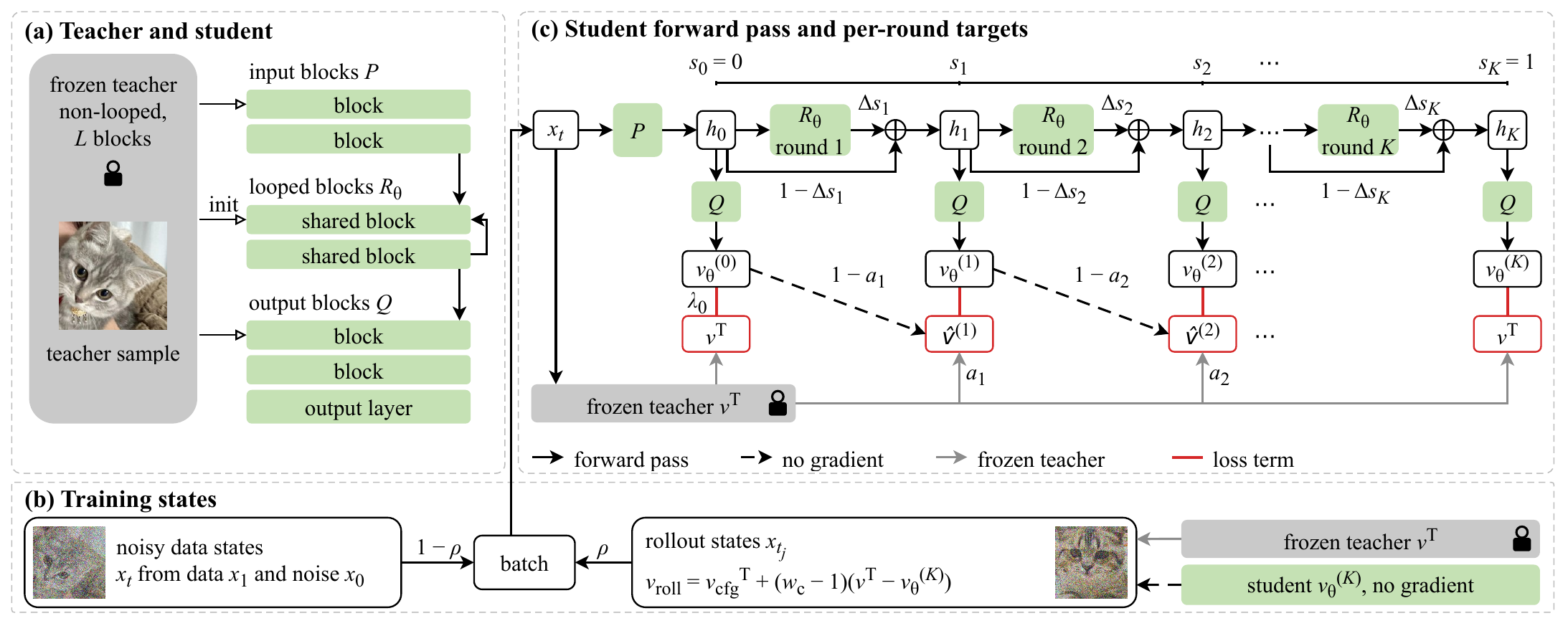}
\caption{Training pipeline of \ours. (a) The looped student is initialized from the frozen non-looped teacher. (b) Each training batch mixes noisy data states with rollout states, which are sampled with $v_{\mathrm{roll}}$ of Eq.~\ref{eq:rollout} and regenerated at every training step; $v^{\mathrm{T}}_{\mathrm{cfg}}$ is the teacher velocity with classifier-free guidance. (c) Each round combines the previous hidden state and the output of the looped blocks with weights $1 - \Delta s_k$ and $\Delta s_k$, where $\Delta s_k = s_k - s_{k-1}$. All rounds share $R_\theta$, conditioned on $s_k$, and all per-loop outputs share $Q$. The output after every round is regressed to its per-round target of Eq.~\ref{eq:target}, which mixes the stop-gradient previous output and $\vT$ with weights $1 - a_k$ and $a_k$, where $a_k = \frac{s_k - s_{k-1}}{1 - s_{k-1}}$, so the last target is $\vT$ itself, and the output $v_\theta^{(0)}$ before the first round is regressed to $\vT$ with weight $\lambda_0$.}
\label{fig:pipeline}
\end{figure*}

In this section, we briefly introduce the background that the method builds on.

\noindent\textbf{Flow Matching.} Flow matching models~\cite{lipman2023flow,liu2023flow,albergo2023building} transport noise $x_0 \sim \mathcal{N}(0, I)$ to data $x_1$ along the path $x_t = t\,x_1 + (1-t)\,x_0$ with $t \in [0,1]$, so that $t=0$ is pure noise and $t=1$ is data. The velocity of this path is $x_1 - x_0$, and a network $v_\theta(x_t, t, c)$ with condition $c$ is trained to regress it:
\begin{equation}
\mathcal{L}_{\mathrm{FM}}(\theta) = \mathbb{E}_{t,\,x_0,\,x_1,\,c}\,\big\|v_\theta(x_t, t, c) - (x_1 - x_0)\big\|_2^2 .
\label{eq:fm}
\end{equation}
Sampling integrates the ODE $\frac{\mathrm{d}x_t}{\mathrm{d}t} = v_\theta(x_t, t, c)$ from $t=0$ to $t=1$ with a numerical solver. 

\noindent\textbf{Guidance.} Classifier-free guidance (CFG)~\cite{ho2022classifier} combines the conditional prediction with the unconditional prediction $v_\theta(x_t, t, \varnothing)$, where $\varnothing$ denotes a null condition. With guidance scale $w$, sampling uses $\tilde v$ in Eq.~\ref{eq:guidance}, and $w = 1$ recovers the conditional prediction. Autoguidance~\cite{karras2024guiding} replaces the unconditional prediction by the prediction $v_{\mathrm{weak}}$ of a weaker version of the same model, for example a smaller or less trained one. It extrapolates from $v_{\mathrm{weak}}$ toward the main model with coefficient $w_{\mathrm{c}} > 1$, omitting the arguments $(x_t, t, c)$ for brevity:
\begin{equation}
\begin{gathered}
\tilde v = v_\theta(x_t, t, \varnothing) + w\,\big(v_\theta(x_t, t, c) - v_\theta(x_t, t, \varnothing)\big), \\
\tilde v = v_{\mathrm{weak}} + w_{\mathrm{c}}\,\big(v_\theta - v_{\mathrm{weak}}\big).
\end{gathered}
\label{eq:guidance}
\end{equation}

\noindent\textbf{Looped Transformers.} A non-looped diffusion transformer stacks $L$ distinct blocks. A looped transformer~\cite{dehghani2019universal,geiping2025scaling} splits them into $L_{\mathrm{in}}$ input blocks $P$, $L_{\mathrm{loop}}$ looped blocks $R_\theta$ whose weights are shared across loops, and $L_{\mathrm{out}}$ output blocks $Q$ followed by the output layer. Given $K$ loops, the model computes
\begin{equation}
\begin{gathered}
h_0 = P(x_t; t, c), \\
\qquad h_k = R_\theta(h_{k-1}; t, c), \;\; k \in \{1, \dots, K\}, \\
v^{(K)} = Q(h_K; t, c).
\end{gathered}
\label{eq:loop}
\end{equation}
The execution depth is $L_{\mathrm{in}} + K L_{\mathrm{loop}} + L_{\mathrm{out}}$ blocks, but the parameters are those of $L_{\mathrm{in}} + L_{\mathrm{loop}} + L_{\mathrm{out}}$ blocks. Eq.~\ref{eq:loop} is a fixed-point iteration toward $h^\star = R_\theta(h^\star; t, c)$. 

\section{Method}
\label{sec:method}

Existing looped diffusion transformers are trained from scratch~\cite{chng2026looped,goyal2026elastic,derakhshani2026lift}. We instead obtain a looped model by distilling a pretrained non-looped diffusion transformer, which reuses the computation already spent on pretraining, and we argue that this is more efficient than training from scratch. This section presents Distill to Loop~(\ours)~and starts from the looped recurrence of Eq.~\ref{eq:loop}. Fig.~\ref{fig:pipeline} shows the training pipeline, and Alg.~\ref{alg:train} and Alg.~\ref{alg:sample} summarize training and sampling.

\begin{figure*}[t]
\centering
\begin{minipage}[t]{0.49\textwidth}\raggedright
\begin{algorithm}[H]
\small
\caption{Training}\label{alg:train}
\begin{algorithmic}[1]
\Repeat
\State $x_t \gets$ data state (Eq.~\ref{eq:fm}) w.p.\ $1-\rho$, else rollout state (Eq.~\ref{eq:rollout})
\State $s_1 < \dots < s_{K-1}$ sorted from $\mathcal{U}(0,1)$, $s_0 = 0$, $s_K = 1$
\State $h_0 = P(x_t; t, c)$, \; $v_\theta^{(0)} = Q(h_0; t, c)$
\For{$k = 1, \dots, K$}
\State $h_k = h_{k-1} + (s_k - s_{k-1})\big(R_\theta(h_{k-1}; s_k, t, c) - h_{k-1}\big)$
\State $v_\theta^{(k)} = Q(h_k; t, c)$
\State $a_k = \frac{s_k - s_{k-1}}{1 - s_{k-1}}$
\State $\hat v^{(k)} = \mathrm{sg}(v_\theta^{(k-1)}) + a_k\big(\vT - \mathrm{sg}(v_\theta^{(k-1)})\big)$
\EndFor
\State $\ell = \frac{1}{K}\sum_{k=1}^{K} \|v_\theta^{(k)} - \hat v^{(k)}\|^2 + \lambda_0 \|v_\theta^{(0)} - \vT\|^2$
\State Take a gradient step on $\nabla_\theta\, \ell$
\Until{converged}
\end{algorithmic}
\end{algorithm}
\end{minipage}\hfill
\begin{minipage}[t]{0.49\textwidth}\raggedright
\begin{algorithm}[H]
\small
\caption{Sampling~(w/o Loop Guidance)}\label{alg:sample}
\begin{algorithmic}[1]
\State $x_{t_0} \sim \mathcal{N}(0, I)$
\For{$j = 0, \dots, N - 1$}
\For{$c' \in \{c, \varnothing\}$}
\State $h_0 = P(x_{t_j}; t_j, c')$
\For{$k = 1, \dots, K$} \Comment{Eq.~\ref{eq:euler}, $s_k = \frac{k}{K}$}
\State $h_k = h_{k-1} + \frac{1}{K}\big(R_\theta(h_{k-1}; s_k, t_j, c') - h_{k-1}\big)$
\EndFor
\State $v(c') = Q(h_K; t_j, c')$
\EndFor
\State $\tilde v = v(\varnothing) + w\big(v(c) - v(\varnothing)\big)$ \Comment{Eq.~\ref{eq:guidance}}
\State $x_{t_{j+1}} = x_{t_j} + (t_{j+1} - t_j)\,\tilde v$
\EndFor
\State \Return $x_{t_N}$
\vspace{6.9pt}%
\end{algorithmic}
\end{algorithm}
\end{minipage}
\end{figure*}

\subsection{Distilling a Looped Transformer}
\label{sec:method:acquisition}

\noindent\textbf{Relaxed Fixed-Point Targets.} A residual fixed-point iteration updates an iterate $z$ by adding a correction $f(z)$:
\begin{equation}
z_{k+1} = z_k + f(z_k), \qquad f(z^\star) = 0 .
\label{eq:fpi}
\end{equation}
The looped recurrence of Eq.~\ref{eq:loop} has this form in the hidden state. Every transformer block is residual, so $R_\theta(h) = h + \Delta_\theta(h)$ and $h_k = h_{k-1} + \Delta_\theta(h_{k-1})$.
Let the teacher be a pretrained non-looped diffusion transformer with velocity $\vT(x_t, t, c)$. The student outputs a velocity after every round, $v_\theta^{(k)} = Q(h_k; t, c)$ for $k \in \{0, \dots, K\}$, where $Q$ is the output module and $v_\theta^{(0)}$ reads $h_0$ before any loop. Distillation should bring these per-loop outputs toward $\vT$, so we ask the per-loop outputs themselves to follow Eq.~\ref{eq:fpi} in velocity space with $\vT$ as the fixed point:
$v_\theta^{(k)} = \mathrm{sg}(v_\theta^{(k-1)}) + f_k(\mathrm{sg}(v_\theta^{(k-1)}))$,
where $\mathrm{sg}$ denotes stop-gradient. The per-loop outputs do not follow this iteration by construction, and the loss below imposes it.
We construct $f_k(v) = a_k\,(\vT - v)$ with $a_k \in (0,1]$, whose only fixed point is $v = \vT$. The target of round $k$ is then
\begin{equation}
\hat{v}^{(k)} = \mathrm{sg}(v_\theta^{(k-1)}) + a_k\,\big(\vT - \mathrm{sg}(v_\theta^{(k-1)})\big),
\label{eq:target}
\end{equation}
so round $k$ closes the fraction $a_k$ of the gap that round $k-1$ left. To let the looped model extend to any number of loops $K$, we index the rounds by continuous depth coordinates $0 = s_0 < s_1 < \dots < s_K = 1$, and round $k$ is conditioned on $s_k$ through a zero-initialized embedding added to the condition of the shared blocks.
The recurrence of Eq.~\ref{eq:loop} takes a full step at every round, so the total update grows with $K$ and different $K$ compute different functions, so we replace it by the relaxed fixed-point iteration of Krasnosel'skii and Mann~\cite{krasnoselskii1955two,mann1953mean}, $h_k = h_{k-1} + \alpha_k\,(R_\theta(h_{k-1}) - h_{k-1})$ with relaxation $\alpha_k \in (0,1]$. The iteration keeps the fixed points of $R_\theta$, and $\alpha_k = 1$ recovers Eq.~\ref{eq:loop}. We set $\alpha_k = s_k - s_{k-1}$. As the steps shrink, the iteration becomes the Euler discretization of the ODE
\begin{equation}
\frac{\mathrm{d}h}{\mathrm{d}s} = R_\theta(h; s, t, c) - h, \qquad s \in [0,1],
\label{eq:depth_ode}
\end{equation}
whose equilibria are the fixed points of $R_\theta$. With the depth condition $s_k$, round $k$ is one Euler step of Eq.~\ref{eq:depth_ode}:
\begin{equation}
h_k = h_{k-1} + (s_k - s_{k-1})\,\big(R_\theta(h_{k-1}; s_k, t, c) - h_{k-1}\big).
\label{eq:euler}
\end{equation}
The relaxations sum to $1$ for every $K$, so every $K$ discretizes Eq.~\ref{eq:depth_ode} from $h_0$ at $s = 0$ to $h_K$ at $s = 1$. The hidden state follows this relaxed iteration, the per-loop outputs are trained to follow the iteration of Eq.~\ref{eq:target} toward $\vT$, and both are indexed by the same coordinates $s_k$.

\noindent\textbf{Training.} During training, $s_1, \dots, s_{K-1}$ are sorted draws from $\mathcal{U}(0,1)$, independent for every sample, so the student sees many depth grids. The coefficient of Eq.~\ref{eq:target} is $a_k = \frac{s_k - s_{k-1}}{1 - s_{k-1}}$, the depth step of round $k$ divided by the depth that remains before it. Round $k$ therefore measures the gap to $\vT$ from the actual output of round $k-1$ and closes the share of that gap that its depth step takes of the remaining depth. If every round meets its target, the output after round $k$ lies at the fraction $s_k$ of the way from $v_\theta^{(0)}$ to $\vT$, and $a_K = 1$ makes the last round target $\vT$ itself. The output $v_\theta^{(0)}$ passes through no round, and the target of round $1$ is built from its stop-gradient copy, so we also regress $v_\theta^{(0)}$ directly on $\vT$ with weight $\lambda_0$, which anchors the start of the output iteration. The full loss averages the squared error over rounds and adds this term, with $x_t$ built from data as in Eq.~\ref{eq:fm}:
\begin{equation}
\begin{aligned}
\mathcal{L}(\theta) = \mathbb{E}_{t,x_0,x_1,c,s}\Big[&\frac{1}{K}\sum_{k=1}^{K}\big\|v_\theta^{(k)}(x_t,t,c,s_k) - \hat{v}^{(k)}\big\|_2^2 \\
&+ \lambda_0\,\big\|v_\theta^{(0)} - \vT\big\|_2^2\Big].
\end{aligned}
\label{eq:loss_fp}
\end{equation}
LiFT~\cite{derakhshani2026lift} trains from scratch toward the flow-matching target $v_{\mathrm{gt}} = x_1 - x_0$ of Eq.~\ref{eq:fm} and anchors every target at the output before any loop, with the target $(1 - s_k)\,\mathrm{sg}(v_\theta^{(0)}) + s_k\,v_{\mathrm{gt}}$ for round $k$. With $v_{\mathrm{gt}}$ in place of $\vT$, Eq.~\ref{eq:target} gives the same target when the output of round $k-1$ lies on this line, so the LiFT target is the special case of Eq.~\ref{eq:target} in which the previous output is replaced by its point on the line. The LiFT target of round $k$ does not depend on where the earlier rounds actually ended. Eq.~\ref{eq:target} starts every round from the actual output of the previous round, so each round also corrects the error that the earlier rounds left.

\noindent\textbf{Inference.} The coefficient $a_k$ appears only in the targets of Eq.~\ref{eq:target}, so it is used only in training. At inference the student solves Eq.~\ref{eq:depth_ode} directly. Discretizing it on the grid $0 = s_0 < \dots < s_K = 1$ gives the Euler step of Eq.~\ref{eq:euler}, whose step is the depth interval $s_k - s_{k-1}$. With $K$ loops the student takes the uniform grid $s_k = \frac{k}{K}$, so every step is $\frac{1}{K}$ and $K$ is chosen at inference without retraining. Because the loops are steps of an ODE, higher-order solvers can replace the Euler step. See App.~\ref{app:depth_solvers} for the depth Heun and DPM++ solvers.
\begin{figure*}[t]
\centering
\includegraphics[width=\textwidth]{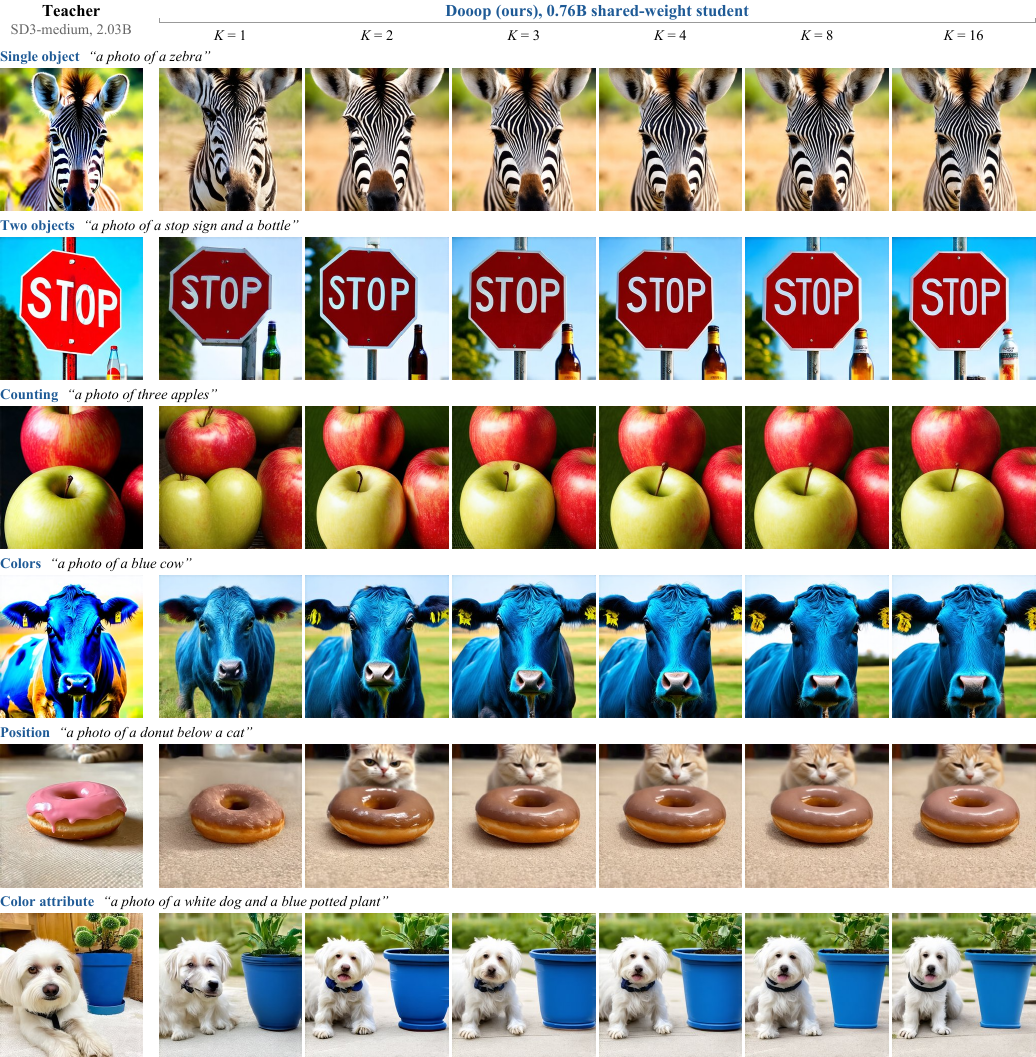}
\caption{Samples of the SD3-medium teacher and of \ours~with $K = 1, 2, 3, 4, 8, 16$ loops on one GenEval prompt from each of the six categories. All images in a row start from the same initial noise.}
\label{fig:qual_sd3}
\end{figure*}

\noindent\textbf{Training on Guided Rollout States.} Eq.~\ref{eq:loss_fp} trains the student only on noisy data states, but at sampling time the student visits the states that its own sampler produces~\cite{ning2023input}. Training on the student's own rollouts does not solve this. While the student is far from the teacher, its states fall outside the distribution the teacher was trained on, and $\vT$ gives no reliable supervision there. We therefore generate the training states with a sampler that combines the teacher and the current student. It follows S$^2$-Guidance~\cite{chen2026stochastic} with the student as the weak model:
\begin{equation}
\begin{aligned}
v_{\mathrm{roll}}(x,t,c) &= \vT(x,t,\varnothing) + w\big(\vT(x,t,c) - \vT(x,t,\varnothing)\big) \\
&\quad + (w_{\mathrm{c}} - 1)\big(\vT(x,t,c) - v_\theta^{(K)}(x,t,c)\big),
\end{aligned}
\label{eq:rollout}
\end{equation}
with $w_{\mathrm{c}} > 1$. For $w = 1$ it is the autoguidance combination~\cite{karras2024guiding} of Eq.~\ref{eq:guidance} with the teacher as the strong model. Our student is the teacher with part of its blocks removed and the rest reused, so it plays the role of the block-dropped sub-network that S$^2$-Guidance extrapolates away from. The student term runs $K$ loops of Eq.~\ref{eq:euler} on the inference grid $s_k = \frac{k}{K}$ with the current weights, without gradient and on the conditional branch only. Sampling with $v_{\mathrm{roll}}$ from noise for a training caption $c$ gives the states $x^{\mathrm{roll}}_t$, which are regenerated at every training step. These states change with the student and stay near the teacher's sampling trajectories. On every state $x^{\mathrm{roll}}_t$ the student is trained with the targets of Eq.~\ref{eq:target}, whose endpoint is the unguided teacher velocity $\vT(x^{\mathrm{roll}}_t, t, c)$. The guidance scales $w$ and $w_{\mathrm{c}}$ only shape the visited states. With $\ell$ the integrand of Eq.~\ref{eq:loss_fp}, a fraction $\rho$ of every batch comes from rollout states and the rest from noisy data states~(ablation see Sec.~\ref{sec:exp:ablation}).:
\begin{equation}
\mathcal{L}(\theta) = (1-\rho)\,\mathbb{E}_{\mathrm{data}}\big[\ell\big] + \rho\,\mathbb{E}_{\mathrm{roll}}\big[\ell\big].
\label{eq:loss_total}
\end{equation}
\begin{table*}[t]
\small
\centering
\caption{Text-to-image generation with SD3-medium as the teacher. Best student result of each part in bold.}
\label{tab:sd3}
\setlength{\tabcolsep}{2pt}
\begin{tabular}{lccccccccccccc}
\toprule
 & & & \multicolumn{7}{c}{GenEval~\cite{ghosh2023geneval}$\uparrow$} & & & & \\
\cmidrule(lr){4-10}
Model & TFLOPs & Rounds & \shortstack{Single\\obj.} & \shortstack{Two\\obj.} & Counting & Colors & Position & \shortstack{Color\\attr.} & Overall & \shortstack{Image-\\Reward\\\cite{xu2023imagereward}$\uparrow$} & \shortstack{Pick-\\Score\\\cite{kirstain2023pick}$\uparrow$} & \shortstack{HPS-\\v2.1\\\cite{wu2023human}$\uparrow$} & \shortstack{Avg.\\(\%)}$\uparrow$ \\
\midrule
SD3-medium~\cite{esser2024scaling} (teacher) & 50.5 & -- & 0.994 & 0.753 & 0.506 & 0.840 & 0.285 & 0.508 & 0.648 & 0.908 & 22.214 & 0.278 & 100.0 \\
\midrule
\rowcolor{gray!15}\multicolumn{14}{l}{\textit{Distill}} \\
\quad \ours~(Ours) & 18.9 & 1 & 0.988 & 0.737 & 0.322 & 0.811 & 0.248 & 0.428 & 0.589 & 0.663 & 21.790 & 0.259 & 88.8 \\
\quad \ours~(Ours) & 29.4 & 2 & \textbf{0.991} & \textbf{0.838} & 0.456 & \textbf{0.856} & \textbf{0.288} & 0.535 & \textbf{0.661} & 0.924 & 22.245 & 0.275 & 100.7 \\
\quad \ours~(Ours) & 39.9 & 3 & 0.984 & 0.813 & 0.466 & 0.843 & 0.270 & 0.550 & 0.654 & \textbf{0.933} & 22.307 & \textbf{0.276} & \textbf{100.9} \\
\quad \ours~(Ours) & 50.5 & 4 & 0.988 & 0.795 & \textbf{0.503} & 0.830 & 0.273 & \textbf{0.565} & 0.659 & 0.919 & \textbf{22.312} & \textbf{0.276} & 100.7 \\
\quad \ours~(Ours) & 92.6 & 8 & 0.984 & 0.803 & 0.475 & 0.832 & 0.245 & 0.548 & 0.648 & 0.889 & 22.278 & 0.275 & 99.3 \\
\quad \ours~(Ours) & 177.0 & 16 & \textbf{0.991} & 0.760 & 0.438 & 0.838 & 0.250 & 0.525 & 0.633 & 0.853 & 22.229 & 0.273 & 97.6 \\
\midrule
\rowcolor{gray!15}\multicolumn{14}{l}{+~\textit{Real data~(Keep training after distill use ground truth)}} \\
\quad \ours~(Ours) & 18.9 & 1 & 0.994 & 0.811 & 0.394 & 0.846 & 0.305 & 0.495 & 0.641 & 0.770 & 21.831 & 0.267 & 94.5 \\
\quad \ours~(Ours) & 29.4 & 2 & \textbf{0.997} & 0.879 & 0.497 & 0.880 & \textbf{0.353} & \textbf{0.635} & 0.707 & 1.021 & 22.282 & 0.281 & 105.8 \\
\quad \ours~(Ours) & 39.9 & 3 & 0.994 & \textbf{0.889} & 0.522 & \textbf{0.886} & 0.343 & 0.608 & 0.707 & \textbf{1.039} & \textbf{22.391} & \textbf{0.284} & \textbf{106.6} \\
\quad \ours~(Ours) & 50.5 & 4 & \textbf{0.997} & 0.884 & \textbf{0.547} & 0.878 & 0.343 & 0.610 & \textbf{0.710} & 1.037 & 22.390 & 0.283 & \textbf{106.6} \\
\bottomrule
\end{tabular}
\end{table*}

\subsection{Guiding with Early Loops}
\label{sec:method:guidance}
Motivated by autoguidance~\cite{karras2024guiding} and S$^2$-Guidance~\cite{chen2026stochastic}, we note that a looped model run with different loop counts naturally forms the strong and weak pair that Eq.~\ref{eq:guidance} requires. The student with $K$ loops is the strong model, and the same student with fewer loops $k < K$ is the weak model. We therefore propose loop guidance, which samples with the velocity
\begin{equation}
\begin{aligned}
v_{\mathrm{lg}}(x,t,c) &= v_\theta^{(K)}(x,t,\varnothing) \\
&+ w\big(v_\theta^{(K)}(x,t,c) - v_\theta^{(K)}(x,t,\varnothing)\big) \\
&+ (w_{\mathrm{l}} - 1)\big(v_\theta^{(K)}(x,t,c) - v_{\mathrm{weak}}(x,t,c)\big),
\end{aligned}
\label{eq:loop_guidance}
\end{equation}
where $v_{\mathrm{weak}}$ is the weak prediction from $k$ loops and $w_{\mathrm{l}}$ is the loop guidance scale. The weak prediction takes one of two forms. It is either the output $v_\theta^{(k)}$ after loop $k$ of the same $K$-loop forward, which costs one extra application of the output module $Q$, or the output $v_\theta^{[k]}$ of a separate forward with $k$ loops. See App.~\ref{app:loop_guidance} for the sampling algorithms of both forms.

\section{Experiments}
\label{sec:exp}
In this section, we validate the effectiveness of \ours~through multiple experiments and benchmarks on SD3-medium and SiT-XL/2, and we show qualitative results (see Fig.~\ref{fig:qual_sd3}). See App.~\ref{app:experiments} and App.~\ref{app:qual} for more experiments and qualitative results.

\begin{table*}[t]
\footnotesize
\centering
\caption{GenEval and ImageReward of the SD3-medium teacher and of \ours~for each time sampler (rows) and each depth solver of the loop with $K$ loops (columns). The italic row gives the GFLOPs of one function evaluation (NFE), so the compute per sample is the number of NFEs times this value. Bold marks the highest value among the \ours~columns in each row for each metric. Darker cells are better within each row and metric.}
\label{tab:samplers_sd3}
\setlength{\tabcolsep}{1.3pt}
\begin{tabular}{lc@{\hspace{3pt}}cc@{\hspace{3pt}}cc@{\hspace{3pt}}cc@{\hspace{3pt}}cc@{\hspace{3pt}}cc@{\hspace{3pt}}cc@{\hspace{3pt}}cc@{\hspace{3pt}}cc}
\toprule
 & & & & \multicolumn{14}{c}{Depth solver of \ours} \\
\cmidrule(lr){5-18}
 & & & & \multicolumn{6}{c}{Euler} & \multicolumn{4}{c}{Heun~\cite{karras2022elucidating}} & \multicolumn{4}{c}{DPM++~\cite{lu2025dpm}} \\
\cmidrule(lr){5-10}\cmidrule(lr){11-14}\cmidrule(lr){15-18}
 & & \multicolumn{2}{c}{Teacher} & \multicolumn{2}{c}{$K{=}2$} & \multicolumn{2}{c}{$K{=}4$} & \multicolumn{2}{c}{$K{=}8$} & \multicolumn{2}{c}{$K{=}2$} & \multicolumn{2}{c}{$K{=}4$} & \multicolumn{2}{c}{$K{=}2$} & \multicolumn{2}{c}{$K{=}4$} \\
\cmidrule(lr){3-4}\cmidrule(lr){5-6}\cmidrule(lr){7-8}\cmidrule(lr){9-10}\cmidrule(lr){11-12}\cmidrule(lr){13-14}\cmidrule(lr){15-16}\cmidrule(lr){17-18}
Sampler & Steps & \shortstack{Gen-\\Eval$\uparrow$} & \shortstack{Image-\\Reward$\uparrow$} & \shortstack{Gen-\\Eval$\uparrow$} & \shortstack{Image-\\Reward$\uparrow$} & \shortstack{Gen-\\Eval$\uparrow$} & \shortstack{Image-\\Reward$\uparrow$} & \shortstack{Gen-\\Eval$\uparrow$} & \shortstack{Image-\\Reward$\uparrow$} & \shortstack{Gen-\\Eval$\uparrow$} & \shortstack{Image-\\Reward$\uparrow$} & \shortstack{Gen-\\Eval$\uparrow$} & \shortstack{Image-\\Reward$\uparrow$} & \shortstack{Gen-\\Eval$\uparrow$} & \shortstack{Image-\\Reward$\uparrow$} & \shortstack{Gen-\\Eval$\uparrow$} & \shortstack{Image-\\Reward$\uparrow$} \\
\midrule
\textit{GFLOPs/NFE} & & \multicolumn{2}{c}{\textit{1802.6}} & \multicolumn{2}{c}{\textit{1049.8}} & \multicolumn{2}{c}{\textit{1802.7}} & \multicolumn{2}{c}{\textit{3308.4}} & \multicolumn{2}{c}{\textit{1426.2}} & \multicolumn{2}{c}{\textit{2931.9}} & \multicolumn{2}{c}{\textit{1049.8}} & \multicolumn{2}{c}{\textit{1802.7}} \\
\midrule
\multirow{3}{*}{Euler} & 10 & 0.535 & 0.449 & \cellcolor{ForestGreen!26}0.595 & \cellcolor{ForestGreen!33}0.656 & \cellcolor{ForestGreen!33}0.605 & \cellcolor{ForestGreen!26}0.650 & \cellcolor{ForestGreen!19}0.591 & \cellcolor{ForestGreen!6}0.585 & \cellcolor{ForestGreen!12}0.587 & \cellcolor{ForestGreen!19}0.618 & 0.568 & 0.544 & \cellcolor{ForestGreen!40}\textbf{0.616} & \cellcolor{ForestGreen!40}\textbf{0.703} & \cellcolor{ForestGreen!6}0.579 & \cellcolor{ForestGreen!12}0.588 \\
 & 25 & 0.626 & 0.882 & \cellcolor{ForestGreen!33}0.660 & \cellcolor{ForestGreen!26}0.911 & \cellcolor{ForestGreen!26}0.658 & \cellcolor{ForestGreen!33}0.935 & \cellcolor{ForestGreen!6}0.638 & \cellcolor{ForestGreen!6}0.897 & 0.633 & \cellcolor{ForestGreen!19}0.909 & \cellcolor{ForestGreen!12}0.644 & 0.869 & \cellcolor{ForestGreen!40}\textbf{0.665} & \cellcolor{ForestGreen!40}\textbf{0.957} & \cellcolor{ForestGreen!19}0.648 & \cellcolor{ForestGreen!12}0.899 \\
 & 50 & 0.650 & 0.991 & \cellcolor{ForestGreen!33}0.666 & \cellcolor{ForestGreen!19}0.960 & \cellcolor{ForestGreen!40}\textbf{0.672} & \cellcolor{ForestGreen!33}0.977 & \cellcolor{ForestGreen!26}0.659 & \cellcolor{ForestGreen!12}0.959 & \cellcolor{ForestGreen!6}0.650 & \cellcolor{ForestGreen!33}0.977 & 0.642 & 0.941 & \cellcolor{ForestGreen!19}0.658 & \cellcolor{ForestGreen!40}\textbf{0.983} & \cellcolor{ForestGreen!12}0.652 & \cellcolor{ForestGreen!6}0.956 \\
\cmidrule(l){2-18}
\multirow{2}{*}{Heun~\cite{karras2022elucidating}} & 10 & 0.559 & 0.714 & \cellcolor{ForestGreen!26}0.618 & \cellcolor{ForestGreen!33}0.855 & \cellcolor{ForestGreen!33}0.625 & \cellcolor{ForestGreen!26}0.849 & \cellcolor{ForestGreen!12}0.614 & \cellcolor{ForestGreen!12}0.813 & \cellcolor{ForestGreen!6}0.606 & \cellcolor{ForestGreen!19}0.821 & 0.601 & 0.776 & \cellcolor{ForestGreen!40}\textbf{0.629} & \cellcolor{ForestGreen!40}\textbf{0.889} & \cellcolor{ForestGreen!19}0.616 & \cellcolor{ForestGreen!6}0.806 \\
 & 25 & 0.654 & 0.950 & \cellcolor{ForestGreen!26}0.662 & \cellcolor{ForestGreen!26}0.966 & \cellcolor{ForestGreen!33}0.666 & \cellcolor{ForestGreen!33}0.970 & \cellcolor{ForestGreen!19}0.657 & \cellcolor{ForestGreen!12}0.947 & 0.637 & 0.928 & \cellcolor{ForestGreen!6}0.643 & \cellcolor{ForestGreen!6}0.933 & \cellcolor{ForestGreen!40}\textbf{0.669} & \cellcolor{ForestGreen!40}\textbf{0.989} & \cellcolor{ForestGreen!12}0.645 & \cellcolor{ForestGreen!19}0.950 \\
\cmidrule(l){2-18}
\multirow{2}{*}{\shortstack[l]{DPM++\\\cite{lu2025dpm}}} & 10 & 0.374 & 0.324 & \cellcolor{ForestGreen!33}0.586 & \cellcolor{ForestGreen!33}0.716 & \cellcolor{ForestGreen!40}\textbf{0.599} & \cellcolor{ForestGreen!26}0.707 & \cellcolor{ForestGreen!19}0.568 & \cellcolor{ForestGreen!12}0.635 & 0.524 & \cellcolor{ForestGreen!19}0.641 & \cellcolor{ForestGreen!12}0.547 & 0.575 & \cellcolor{ForestGreen!26}0.572 & \cellcolor{ForestGreen!40}\textbf{0.738} & \cellcolor{ForestGreen!6}0.546 & \cellcolor{ForestGreen!6}0.617 \\
 & 25 & 0.600 & 0.893 & \cellcolor{ForestGreen!19}0.639 & \cellcolor{ForestGreen!26}0.951 & \cellcolor{ForestGreen!40}\textbf{0.657} & \cellcolor{ForestGreen!33}0.973 & \cellcolor{ForestGreen!26}0.644 & \cellcolor{ForestGreen!12}0.930 & 0.631 & \cellcolor{ForestGreen!19}0.936 & \cellcolor{ForestGreen!6}0.632 & 0.896 & \cellcolor{ForestGreen!33}0.652 & \cellcolor{ForestGreen!40}\textbf{0.987} & \cellcolor{ForestGreen!12}0.636 & \cellcolor{ForestGreen!6}0.927 \\
\bottomrule
\end{tabular}
\end{table*}

\begin{figure*}[t]
\centering
\includegraphics[width=1\textwidth]{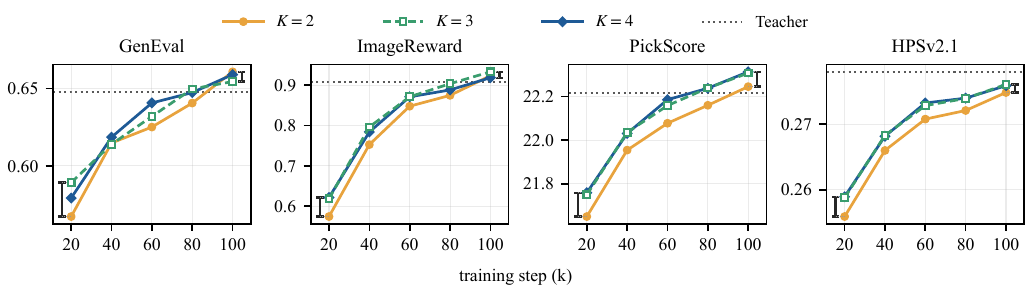}
\caption{GenEval, ImageReward, PickScore and HPSv2.1 of \ours~on SD3-medium with $K = 2, 3, 4$ loops during training. The dotted line marks the teacher.}
\label{fig:kcurve_sd3}
\end{figure*}

\begin{table}[t]
\footnotesize
\centering
\caption{Comparison with looped models trained from scratch and cumulative ablation of \ours~on ImageNet $256\times256$ class-conditional generation. Best result in bold.}
\label{tab:main}
\setlength{\tabcolsep}{1.8pt}
\begin{tabular}{lcccc}
\toprule
Method & Params & \shortstack{Train\\samples} & FID$\downarrow$ & IS$\uparrow$ \\
\midrule
ELT $4N{\times}8L$~\cite{goyal2026elastic} & 271M & 256M & 3.96 & -- \\
\midrule
Looped (from scratch) & 246M & 77.3M & 12.573 & 72.560 \\
\quad + init.\ from teacher & 246M & 77.3M & 6.313 & 147.937 \\
\quad + distillation (\ours) & 246M & 77.3M & \textbf{3.785} & \textbf{221.573} \\
\bottomrule
\end{tabular}
\end{table}

\begin{table}[t]
\small
\centering
\caption{Loop guidance on the CDM-distilled four-step SD3 student. The teacher is the four-step CDM model. The weak model is a separate forward with $k$ loops or the output after loop $k$ of the same $K{=}4$ forward. Best student result in bold.}
\label{tab:loop_guidance}
\setlength{\tabcolsep}{1.6pt}
\begin{tabular}{lcccccccc}
\toprule
Setting & $w_{\mathrm{l}}$ & NFE & TFLOPs & \shortstack{Gen-\\Eval$\uparrow$} & \shortstack{Image-\\Reward$\uparrow$} & \shortstack{Pick-\\Score$\uparrow$} & \shortstack{HPS-\\v2.1$\uparrow$} & Avg.$\uparrow$ \\
\midrule
Teacher & -- & 4 & 7.21 & 0.651 & 1.026 & 22.822 & 0.298 & 100.0 \\
Student & 1 & 4 & 7.21 & 0.592 & 0.739 & 22.381 & 0.273 & 88.2 \\
\midrule
\rowcolor{gray!15}\multicolumn{9}{l}{\textit{Weak model~(separate $k$-loop forward)}} \\
\quad $k{=}1$ & 2 & 8 & 9.90 & 0.639 & 0.941 & 22.568 & 0.290 & 96.5 \\
\quad $k{=}1$ & 3 & 8 & 9.90 & 0.636 & 0.975 & 22.416 & 0.295 & 97.5 \\
\quad $k{=}2$ & 3 & 8 & 11.41 & 0.627 & 0.907 & 22.540 & 0.288 & 95.0 \\
\quad $k{=}2$ & 4 & 8 & 11.41 & 0.631 & 0.935 & 22.488 & 0.291 & 96.1 \\
\midrule
\rowcolor{gray!15}\multicolumn{9}{l}{\textit{Weak model~(loop-$k$ output of the same forward)}} \\
\quad $k{=}1$ & 2 & 4 & 7.79 & \textbf{0.643} & 1.002 & 22.575 & 0.296 & 98.7 \\
\quad $k{=}1$ & 3 & 4 & 7.79 & 0.633 & \textbf{1.029} & 22.248 & 0.293 & 98.3 \\
\quad $k{=}2$ & 2 & 4 & 7.79 & 0.634 & 0.971 & \textbf{22.613} & 0.294 & 97.4 \\
\quad $k{=}2$ & 3 & 4 & 7.79 & \textbf{0.643} & 1.025 & 22.454 & \textbf{0.297} & \textbf{99.2} \\
\bottomrule
\end{tabular}
\end{table}

\begin{table}[t]
\small
\centering
\caption{Analysis of rollout ratio in each training batch, at 100k training steps and $K{=}4$. Best result in each column in bold.}
\label{tab:ratio}
\setlength{\tabcolsep}{2pt}
\begin{tabular}{lcccccc}
\toprule
 & \multicolumn{2}{c}{ImageNet (SiT)} & \multicolumn{4}{c}{Text-to-image (SD3)} \\
\cmidrule(lr){2-3}\cmidrule(lr){4-7}
\shortstack{Data :\\rollout} & gFID$\downarrow$ & FD$\downarrow$ & GenEval$\uparrow$ & \shortstack{Image-\\Reward$\uparrow$} & \shortstack{Pick-\\Score$\uparrow$} & HPSv2.1$\uparrow$ \\
\midrule
$1:0$ & \textbf{8.917} & \textbf{305.729} & 0.658 & 0.901 & \textbf{22.354} & \textbf{0.276} \\
$2:1$ & 9.275 & 321.364 & \textbf{0.659} & \textbf{0.919} & 22.312 & \textbf{0.276} \\
$1:2$ & 9.559 & 323.307 & 0.655 & 0.910 & 22.293 & \textbf{0.276} \\
$0:1$ & 11.509 & 359.354 & 0.641 & 0.903 & 22.233 & 0.274 \\
\bottomrule
\end{tabular}
\end{table}

\subsection{Setup}
\label{sec:exp:setup}

\noindent\textbf{Models.} The teacher is the text-to-image model SD3-medium~\cite{esser2024scaling}, with 24 MMDiT blocks and 2.03B parameters. Its student has 2 input, 5 looped and 2 output blocks with 0.756B parameters and is trained with $K = 4$ rounds. The looped blocks use QK normalization and a soft variant of the exclusive self-attention that Looped-DiT~\cite{chng2026looped} introduced for looped diffusion transformers. See App.~\ref{app:setup_other} for the setup of the ImageNet experiments.

\noindent\textbf{Training.} The student trains on the COCO 2017~\cite{lin2014microsoft} training captions and images at $512\times512$. The target endpoint is the conditional velocity $\vT$ of the teacher without guidance at the same $(x_t, t, c)$. Captions are replaced by the empty prompt with probability $0.1$ for the teacher and the student alike, so the student also learns the unconditional velocity. We draw $t \sim \mathcal{U}(0,1)$, the $K-1$ inner depth coordinates are sorted draws from $\mathcal{U}(0,1)$ for every sample, and $\lambda_0 = 0.1$. One third of every batch comes from rollout states ($\rho = \frac{1}{3}$), with the teacher CFG scale in Eq.~\ref{eq:rollout} set to $4.5$ and $w_{\mathrm{c}} = 2$. Training uses AdamW and an exponential moving average of the weights on 2 NVIDIA H200 GPUs. See App.~\ref{app:impl} for the remaining hyperparameters.

\noindent\textbf{Evaluation.} We use GenEval~\cite{ghosh2023geneval} on its 553 prompts with 4 images each at $512\times512$, 28 sampling steps and CFG 7. ImageReward~\cite{xu2023imagereward}, PickScore~\cite{kirstain2023pick} and HPSv2.1~\cite{wu2023human} score the same images. Avg. in the tables is the mean of the four metrics, each divided by the teacher's value, so the teacher is 100. FLOPs in the tables are the total for generating one image, count two operations per multiply-add and exclude the extra cost of CFG.

\subsection{Main Results}
\label{sec:exp:main}

\noindent\textbf{Qualitative Results.} Fig.~\ref{fig:qual_sd3} shows samples of the teacher and of \ours~with different $K$ from the same initial noise. With fewer than four loops, from $K = 2$ on, the samples are already comparable to the teacher's, and in the position example they place the cat above the donut, which the teacher's sample misses. As the loops continue to $K = 8$ and $K = 16$, the samples stay stable and do not degrade. See App.~\ref{app:qual} for more samples.

\noindent\textbf{Quantitative Results.} Tab.~\ref{tab:sd3} compares the SD3-medium teacher with \ours~at different loop counts on GenEval, ImageReward, PickScore and HPSv2.1, together with the compute of each model. Its upper part reports the student distilled for 100k steps, run with $K = 1$ to $K = 16$ loops from one model. At $K = 2$ the student already reaches the teacher, with GenEval 0.661 against 0.648 and an average of 100.7 over the four metrics, at 29.4 TFLOPs, about 58\% of the teacher's 50.5 TFLOPs. Fig.~\ref{fig:kcurve_sd3} shows the training curves at $K = 2, 3, 4$, where the gap between fewer and more loops narrows as training proceeds, so the student gradually converges at earlier loops. The lower part reports the same student after 20k more steps that use real data as the target in place of the teacher, described in App.~\ref{app:realdata}. At every loop count from $K = 1$ to $K = 4$ it is higher than the distilled student on all four metrics, and at $K = 2$ it reaches GenEval 0.707 and an average of 105.8. \ours~therefore trains a looped model not only by distillation but also with real data as the target.

\subsection{Adjustable Depth at Test Time}
\label{sec:exp:depth}

\noindent\textbf{Depth Solvers.} To test whether the higher-order solvers of diffusion sampling also work in depth, we run the student with each depth solver under each time sampler. Tab.~\ref{tab:samplers_sd3} crosses the time sampler, Euler, Heun~\cite{karras2022elucidating} or DPM++~\cite{lu2022dpm,lu2025dpm}, with the depth solver of the loops. With 10 sampling steps, every student column is above the teacher on GenEval and ImageReward under all three time samplers. The depth DPM++ solver at $K = 2$ gives the best student result in the Euler-10, Euler-25, Heun-10 and Heun-25 rows at about 58\% of the teacher's compute per network evaluation, for example GenEval 0.629 against 0.559 with the Heun sampler at 10 steps. The depth solver is therefore a second test-time choice next to the time sampler. See App.~\ref{app:samplers} for PickScore, HPSv2.1 and the ImageNet counterpart, and App.~\ref{app:depth_solvers} for the definitions of the depth solvers. See App.~\ref{app:adaptive} for adaptive loop counts set for each step.

\subsection{Loop Guidance}
\label{sec:exp:guidance}

To verify that the proposed loop guidance is effective, we apply it to a looped student distilled from the four-step SD3-medium model of CDM~\cite{liu2026continuous}, which is also the teacher in this experiment. Tab.~\ref{tab:loop_guidance} reports the results at CFG 1. Taking the weak prediction from the output after loop $k$ of the same forward raises the average over the four metrics from 88.2 to 99.2 against the teacher's 100, while the compute grows only from 7.21 to 7.79 TFLOPs per image. A separate weak forward doubles the network evaluations and reaches averages of 95.0 to 97.5. Loop guidance therefore brings the student close to the teacher at little extra compute. See App.~\ref{app:loop_guidance} for the analysis of loop guidance on the full-step SD3 student.

\subsection{Ablation Studies and Analyses}
\label{sec:exp:ablation}

\noindent\textbf{Analysis on Rollout Ratio.} To test whether the proposed training on guided rollout states is effective, we vary the ratio of data states to rollout states in each training batch. Tab.~\ref{tab:ratio} reports the results on ImageNet and SD3-medium at 100k steps with $K = 4$. On ImageNet, training on data states only gives the lowest gFID of 8.917, and the gFID rises as the share of rollout states grows. On SD3-medium, a 2:1 mix of data and rollout states gives the highest GenEval of 0.659 and ImageReward of 0.919.

\noindent\textbf{Comparison and ablation.} Tab.~\ref{tab:main} compares \ours~with looped models trained from scratch on ImageNet with SiT-XL/2 as the teacher and adds the components of \ours~one at a time. With 77.3M training samples for each of our rows, initializing the looped model from the teacher lowers the FID from 12.573 to 6.313, and distillation lowers it further to 3.785, while IS improves in the same order. Both the initialization and the distillation therefore contribute to the result. \ours~also outperforms ELT with fewer parameters and fewer training samples.

\section{Conclusion}
\label{sec:conclusion}

Existing looped diffusion transformers are trained from scratch, and they either need a large depth to reach non-looped quality or degrade when they run more loops than in training. We propose Distill to Loop~(\ours), which casts distillation as a fixed-point iteration toward a pretrained teacher and treats each loop as one step of an ordinary differential equation in depth. One student therefore runs any number of loops without retraining, uses higher-order solvers of diffusion sampling along depth, and guides itself with the outputs of its earlier loops. On SD3-medium, two loops already reach the teacher at about 58\% of its compute. Pretrained diffusion transformers can thus become models whose depth is a test-time choice, like sampling steps.
\newpage
\section*{\textsc{LLM Usage Statement}}

In this work, generative AI tools were used for prose polishing and coding
assistance during manuscript and codebase preparation. The research ideas,
experimental design, and result analysis were carried out by the authors.
The authors reviewed all AI-assisted text and code and take responsibility
for the final content of this work, including text, claims, and artifacts
produced with the aid of generative AI.

\FloatBarrier
{
    \small
    \bibliographystyle{ieeenat_fullname}
    \bibliography{main}
}

\clearpage

\setcounter{page}{1}
\maketitlesupplementary

\appendix
\setcounter{figure}{0}
\setcounter{table}{0}
\renewcommand{\thefigure}{S\arabic{figure}}
\renewcommand{\thetable}{S\arabic{table}}
\renewcommand{\theHfigure}{S\arabic{figure}}
\renewcommand{\theHtable}{S\arabic{table}}
\renewcommand{\topfraction}{0.95}
\renewcommand{\bottomfraction}{0.9}
\renewcommand{\textfraction}{0.05}
\renewcommand{\floatpagefraction}{0.5}
\renewcommand{\dbltopfraction}{0.95}
\renewcommand{\dblfloatpagefraction}{0.5}
\setcounter{topnumber}{4}
\setcounter{bottomnumber}{2}
\setcounter{totalnumber}{6}
\setcounter{dbltopnumber}{4}

\section*{Appendix Contents}
\begin{itemize}\setlength{\itemsep}{2pt}
  \item Appendix~\ref{app:related}: Additional related work \dotfill \pageref{app:related}
  \item Appendix~\ref{app:impl_details}: Implementation details \dotfill \pageref{app:impl_details}
  \begin{itemize}\setlength{\itemsep}{1pt}
    \item Appendix~\ref{app:setup_other}: Setup of the ImageNet experiments \dotfill \pageref{app:setup_other}
    \item Appendix~\ref{app:impl}: Training hyperparameters \dotfill \pageref{app:impl}
    \item Appendix~\ref{app:depth_solvers}: Higher-order depth solvers \dotfill \pageref{app:depth_solvers}
  \end{itemize}
  \item Appendix~\ref{app:experiments}: Additional experiments \dotfill \pageref{app:experiments}
  \begin{itemize}\setlength{\itemsep}{1pt}
    \item Appendix~\ref{app:loop_guidance}: Loop guidance \dotfill \pageref{app:loop_guidance}
    \item Appendix~\ref{app:samplers}: Test-time samplers \dotfill \pageref{app:samplers}
    \item Appendix~\ref{app:adaptive}: Adaptive loop counts \dotfill \pageref{app:adaptive}
    \item Appendix~\ref{app:realdata}: Continued training on real data after distillation \dotfill \pageref{app:realdata}
    \item Appendix~\ref{app:zimage}: Results on Z-Image \dotfill \pageref{app:zimage}
  \end{itemize}
  \item Appendix~\ref{app:qual}: Additional qualitative results \dotfill \pageref{app:qual}
  \item Appendix~\ref{app:future}: Limitations and future work \dotfill \pageref{app:future}
\end{itemize}

\section{Additional related work}
\label{app:related}

\noindent\textbf{Efficient and adaptive inference for diffusion models.} Progressive distillation~\cite{salimans2022progressive}, consistency models~\cite{song2023consistency}, distribution matching distillation~\cite{yin2024one,yin2024improved} and CausVid~\cite{yin2025slow} lower the cost of sampling by cutting the number of steps, and they leave the cost of each step unchanged. A direct way to reduce the cost of each step is to compress the depth of a pretrained diffusion model and recover quality by distillation. BK-SDM~\cite{kim2024bksdm} removes blocks of the Stable Diffusion U-Net and distills from the original model, and TinyFusion~\cite{fang2025tinyfusion} learns which layers of DiT and SiT to prune before recovering the pruned model with distillation. The students of both methods keep distinct, unshared layers. Another group adapts the computation to the sample. AdaDiff~\cite{zhang2025adadiff} chooses the number of sampling steps for each prompt, AdaptiveDiffusion~\cite{ye2024training} and ADSC~\cite{binici2026efficient} skip steps online without training, and a second AdaDiff~\cite{tang2024adadiff} exits early from the layers inside a denoising step. DyDiT~\cite{zhao2025dynamic} varies the network width with the timestep and lets some tokens skip the MLP blocks, APTP~\cite{ganjdanesh2025not} assigns each prompt a pruned expert, and Looped-DiT~\cite{chng2026looped} stops its loop early with a learned gate. All of these methods only remove computation from the base model. \ours~sets the loop count at inference and can also run more loops than in training.

\section{Implementation details}
\label{app:impl_details}

\subsection{Setup of the ImageNet experiments}
\label{app:setup_other}
\begin{table*}[t]
\small
\centering
\caption{Training hyperparameters of the three students.}
\label{tab:hparams}
\setlength{\tabcolsep}{4pt}
\begin{tabular}{lccc}
\toprule
 & SiT-XL/2 & SD3-medium & Z-Image \\
\midrule
Teacher blocks / parameters & 28 / 675M & 24 / 2.03B & 30 / 6.15B \\
Resolution & $256$ & $512$ & $512$ \\
Input / looped / output blocks & 2 / 6 / 2 & 2 / 5 / 2 & 3 / 6 / 3 \\
Trainable parameters & 246.2M & 0.756B & 2.90B \\
Training loops $K$ & 4 & 4 & 4 \\
Training data & ImageNet-1k train & COCO 2017 train & COCO 2017 train \\
Condition dropout & 0.1 & 0.1 & 0.1 \\
\midrule
Optimizer & \multicolumn{3}{c}{AdamW, $\beta = (0.9, 0.999)$, no weight decay, gradient clipping at 1.0} \\
Learning rate (constant) & $2.65\times10^{-4}$ & $1.732\times10^{-4}$ & $1.414\times10^{-4}$ \\
Warm-up steps & 500 & 500 & 500 \\
Samples per step & 336 & 144 & 96 \\
Training steps & 230k & 100k & 20k \\
EMA decay & 0.9995 & 0.9995 & 0.9995 \\
Precision (student / teacher) & \multicolumn{3}{c}{bf16 autocast with fp32 master weights / bf16} \\
\midrule
Rollout fraction $\rho$ & 0 & $\frac{1}{3}$ & $\frac{1}{3}$ \\
Rollout Euler steps & -- & 8 & 8 \\
Rollout teacher CFG $w$ & -- & 4.5 & 5.0 \\
$w_{\mathrm{c}}$ & -- & 2 & 2 \\
$\lambda_0$ & 0.1 & 0.1 & 0.1 \\
QK-norm and soft xsa in looped blocks & yes & yes & no \\
\bottomrule
\end{tabular}
\end{table*}

\noindent\textbf{ImageNet.} The teacher is the class-conditional SiT-XL/2~\cite{ma2024sit} on ImageNet~\cite{deng2009imagenet} at $256\times256$, which has 28 blocks and 675M parameters and operates in the latent space of the SD VAE. Its student has 2 input blocks, 6 looped blocks and 2 output blocks, with 246.2M trainable parameters, of which 143.4M lie in the looped blocks, and is trained with $K = 4$ rounds. The input blocks are initialized from teacher blocks 1 and 2, the looped blocks from teacher blocks 3, 7, 11, 15, 19 and 23 (stride 4), and the output blocks from teacher blocks 27 and 28. The looped blocks use QK normalization and a soft variant of the exclusive self-attention that Looped-DiT~\cite{chng2026looped} introduced for looped diffusion transformers. The student trains on the ImageNet training set. Class labels are dropped with probability $0.1$ for the teacher and the student alike, so the student also learns the unconditional velocity. The student trains on noisy data states only ($\rho = 0$) for 230k steps. For evaluation, gFID is the FID against the ADM reference batch of ImageNet~\cite{dhariwal2021diffusion}, and FD-DINOv2 is the Fr\'echet distance in DINOv2 feature space~\cite{stein2023exposing} against 50,000 ImageNet training images. Unless stated otherwise, both use 10,000 samples (10 per class), the Euler sampler with 50 steps, and the uniform depth grid $s_k = \frac{k}{K}$.

\subsection{Training hyperparameters}
\label{app:impl}

Tab.~\ref{tab:hparams} lists the training hyperparameters of the three students. All students draw the inner depth coordinates as sorted uniform draws and report EMA weights. The SiT student first trains for 30k steps, then continues for 200k steps from that checkpoint, including weights, optimizer state and EMA, with no warm-up. Of the 144 samples per step of the SD3 student, 96 are data states and 48 are rollout states.

\subsection{Higher-order depth solvers}
\label{app:depth_solvers}

We write $\Delta_k = s_k - s_{k-1}$ and $R_k = R_\theta(h_{k-1}; s_k)$, and we omit the timestep $t$ and the condition $c$ from the notation. Eq.~\ref{eq:euler} is then the Euler step $h_k = h_{k-1} + \Delta_k (R_k - h_{k-1})$. The depth Heun solver follows the Heun sampler of EDM~\cite{karras2022elucidating}. For $k = 1, \dots, K-1$ it first takes the Euler step as a predictor $\tilde h$ and then averages the slopes at both ends of the step in the corrector,
\begin{equation}
\begin{aligned}
\tilde h &= h_{k-1} + \Delta_k (R_k - h_{k-1}), \\
h_k &= h_{k-1} + \frac{\Delta_k}{2}\Big[(R_k - h_{k-1}) + \big(R_\theta(\tilde h; s_k + \Delta_k) - \tilde h\big)\Big],
\end{aligned}
\label{eq:depth_heun}
\end{equation}
and the last step $k = K$ is the Euler step of Eq.~\ref{eq:euler}. Every step except the last evaluates the shared blocks twice, so one forward pass runs them $2K - 1$ times.

The depth DPM++ solver is a multistep solver that reuses the evaluation of the previous step. Its first step is the Euler step. For $k \geq 2$ it replaces $R_k$ by an extrapolation from $R_k$ and $R_{k-1}$,
\begin{equation}
\begin{aligned}
\hat R_k &= \Big(1 + \frac{1}{2 r_k}\Big) R_k - \frac{1}{2 r_k} R_{k-1}, \\
h_k &= h_{k-1} + \Delta_k (\hat R_k - h_{k-1}),
\end{aligned}
\label{eq:depth_dpm}
\end{equation}
with the step ratio $r_k = \frac{\Delta_{k-1}}{\Delta_k}$. Each step evaluates the shared blocks once, so one forward pass runs them $K$ times, the same as the Euler solver. This number of passes of the shared blocks, $K$ for Euler and DPM++ and $2K - 1$ for Heun, sets the GFLOPs of one function evaluation in the sampler tables. On the uniform grid $s_k = \frac{k}{K}$ we have $r_k = 1$, and Eq.~\ref{eq:depth_dpm} equals the second-order multistep DPM-Solver++ (2M)~\cite{lu2025dpm} applied to Eq.~\ref{eq:depth_ode}, with $R_\theta$ as the data prediction and the step $\lambda = -\ln(1 - \frac{1}{K})$. The exponential weights of that solver are $e^{-\lambda} = 1 - \frac{1}{K}$ and $1 - e^{-\lambda} = \frac{1}{K}$, which are the Euler weights $1 - \Delta_k$ and $\Delta_k$ of Eq.~\ref{eq:euler}. DPM-Solver~\cite{lu2022dpm} introduced this family of solvers.

\section{Additional experiments}
\label{app:experiments}

\begin{figure*}[t]
\centering
\includegraphics[width=\textwidth]{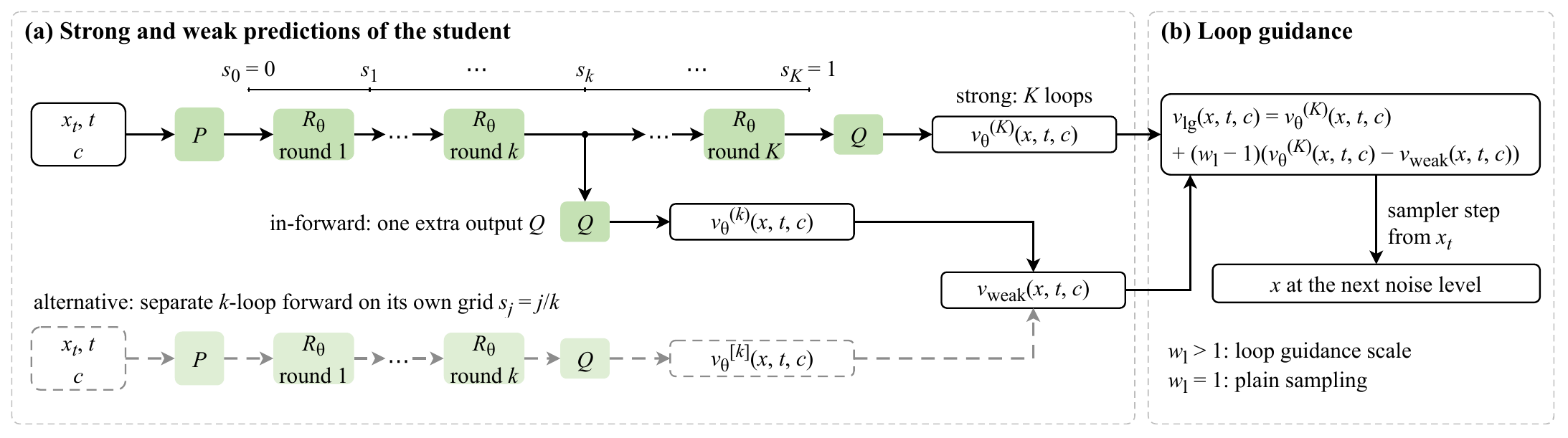}
\caption{Loop guidance at one sampling step of the four-step \ours~student, which is sampled without CFG, so only the conditional branch is used. (a) The strong prediction $v_\theta^{(K)}(x,t,c)$ is the output of $Q$ after round $K$ of the $K$-loop forward. The weak prediction $v_{\mathrm{weak}}(x,t,c)$ is the output $v_\theta^{(k)}$ after round $k$ of the same forward, which costs one extra application of $Q$. The dashed path shows the alternative, a separate forward with $k$ loops on its own grid $s_j = \frac{j}{k}$ that outputs $v_\theta^{[k]}$. (b) For $w = 1$, Eq.~\ref{eq:loop_guidance} reduces to $v_{\mathrm{lg}} = v_\theta^{(K)}(x,t,c) + (w_{\mathrm{l}} - 1)\big(v_\theta^{(K)}(x,t,c) - v_{\mathrm{weak}}(x,t,c)\big)$ with the loop guidance scale $w_{\mathrm{l}} > 1$, and the sampler moves $x_t$ to the next noise level with $v_{\mathrm{lg}}$; $w_{\mathrm{l}} = 1$ recovers plain sampling. All rounds share $R_\theta$ and all per-loop outputs share $Q$.}
\label{fig:loop_guidance}
\end{figure*}

Fig.~\ref{fig:loop_guidance} illustrates the loop guidance of Sec.~\ref{sec:method:guidance} at one sampling step.

Alg.~\ref{alg:sample_lg} and Alg.~\ref{alg:sample_lg_sep} give sampling with loop guidance. In Alg.~\ref{alg:sample_lg} the weak model is the output after loop $m < K$ of the same forward, and in Alg.~\ref{alg:sample_lg_sep} it is a separate forward with $m$ loops.
\begin{figure*}[t]
\centering
\begin{minipage}[t]{0.49\textwidth}\raggedright
\begin{algorithm}[H]
\small
\caption{Sampling~(w/ Loop Guidance, output after loop $m$)}\label{alg:sample_lg}
\begin{algorithmic}[1]
\State $x_{t_0} \sim \mathcal{N}(0, I)$
\For{$j = 0, \dots, N - 1$}
\For{$c' \in \{c, \varnothing\}$}
\State $h_0 = P(x_{t_j}; t_j, c')$
\For{$k = 1, \dots, K$} \Comment{Eq.~\ref{eq:euler}, $s_k = \frac{k}{K}$}
\State $h_k = h_{k-1} + \frac{1}{K}\big(R_\theta(h_{k-1}; s_k, t_j, c') - h_{k-1}\big)$
\EndFor
\State $v(c') = Q(h_K; t_j, c')$
\If{$c' = c$}
\State $v_{\mathrm{weak}} = Q(h_m; t_j, c)$ \Comment{$m < K$}
\EndIf
\EndFor
\State $\tilde v = v(\varnothing) + w\big(v(c) - v(\varnothing)\big)$
\State \hspace{1.2em}$+ (w_{\mathrm{l}} - 1)\big(v(c) - v_{\mathrm{weak}}\big)$ \Comment{Eq.~\ref{eq:loop_guidance}}
\State $x_{t_{j+1}} = x_{t_j} + (t_{j+1} - t_j)\,\tilde v$
\EndFor
\State \Return $x_{t_N}$
\end{algorithmic}
\end{algorithm}
\end{minipage}\hfill
\begin{minipage}[t]{0.49\textwidth}\raggedright
\begin{algorithm}[H]
\small
\caption{Sampling~(w/ Loop Guidance, separate $m$-loop forward)}\label{alg:sample_lg_sep}
\begin{algorithmic}[1]
\State $x_{t_0} \sim \mathcal{N}(0, I)$
\For{$j = 0, \dots, N - 1$}
\For{$c' \in \{c, \varnothing\}$}
\State $h_0 = P(x_{t_j}; t_j, c')$
\For{$k = 1, \dots, K$} \Comment{Eq.~\ref{eq:euler}, $s_k = \frac{k}{K}$}
\State $h_k = h_{k-1} + \frac{1}{K}\big(R_\theta(h_{k-1}; s_k, t_j, c') - h_{k-1}\big)$
\EndFor
\State $v(c') = Q(h_K; t_j, c')$
\EndFor
\State $g_0 = P(x_{t_j}; t_j, c)$ \Comment{weak forward, $m < K$}
\For{$i = 1, \dots, m$} \Comment{$s_i = \frac{i}{m}$}
\State $g_i = g_{i-1} + \frac{1}{m}\big(R_\theta(g_{i-1}; s_i, t_j, c) - g_{i-1}\big)$
\EndFor
\State $v_{\mathrm{weak}} = Q(g_m; t_j, c)$
\State $\tilde v = v(\varnothing) + w\big(v(c) - v(\varnothing)\big)$
\State \hspace{1.2em}$+ (w_{\mathrm{l}} - 1)\big(v(c) - v_{\mathrm{weak}}\big)$ \Comment{Eq.~\ref{eq:loop_guidance}}
\State $x_{t_{j+1}} = x_{t_j} + (t_{j+1} - t_j)\,\tilde v$
\EndFor
\State \Return $x_{t_N}$
\end{algorithmic}
\end{algorithm}
\end{minipage}
\end{figure*}

\subsection{Loop guidance}
\label{app:loop_guidance}
To test whether loop guidance adds to CFG on the full-step SD3 student, Tab.~\ref{tab:cfg_lg} compares raising the CFG scale with adding loop guidance ($w_{\mathrm{l}}$) on top of CFG, where the weak model is the output after loop 2 of the same $K{=}4$ forward. At CFG 1, every metric rises with $w_{\mathrm{l}}$, with GenEval from 0.201 at $w_{\mathrm{l}} = 1$ to 0.335 at $w_{\mathrm{l}} = 3$. At CFG 4.5 and above, loop guidance moves the metrics in the same direction as a larger CFG scale, and raising the CFG scale from 4.5 to 7 without loop guidance scores higher on all four metrics than any tested loop guidance strength at CFG 4.5. Loop guidance is therefore most useful for few-step models sampled without CFG. On the four-step student of Tab.~\ref{tab:loop_guidance}, taking the weak model from the same forward adds only one output computation (7.21 to 7.79 TFLOPs), and GenEval rises from 0.592 to 0.643. The TFLOPs of Tab.~\ref{tab:cfg_lg} follow the convention of Tab.~\ref{tab:sd3}: 28 steps per image, CFG doubling not counted, and loop guidance applied to the conditional branch only, at the cost of one extra output.
\begin{table}[t]
\small
\centering
\caption{The SD3 student (the Distill checkpoint of Tab.~\ref{tab:sd3}, $K{=}4$) at three CFG scales, with and without loop guidance whose weak model is the output after loop 2 of the same forward. Bold marks the best value within each CFG scale.}
\label{tab:cfg_lg}
\setlength{\tabcolsep}{1.6pt}
\begin{tabular}{lcccccccc}
\toprule
Setting & $w$ & $w_{\mathrm{l}}$ & TFLOPs & \shortstack{Gen-\\Eval$\uparrow$} & \shortstack{Image-\\Reward$\uparrow$} & \shortstack{Pick-\\Score$\uparrow$} & \shortstack{HPS-\\v2.1$\uparrow$} & \shortstack{Avg.\\(\%)$\uparrow$} \\
\midrule
Teacher & 7 & -- & 50.5 & 0.648 & 0.908 & 22.214 & 0.278 & 100.0 \\
\midrule
\ours & 1 & 1 & 50.5 & 0.201 & $-$1.160 & 19.584 & 0.198 & 15.6 \\
\ours & 1 & 1.5 & 54.5 & 0.260 & $-$0.802 & 20.026 & 0.214 & 29.7 \\
\ours & 1 & 2 & 54.5 & 0.303 & $-$0.568 & 20.349 & 0.226 & 39.2 \\
\ours & 1 & 3 & 54.5 & \textbf{0.335} & \textbf{$-$0.360} & \textbf{20.599} & \textbf{0.236} & \textbf{47.4} \\
\midrule
\ours & 4.5 & 1 & 50.5 & \textbf{0.632} & 0.788 & 22.074 & 0.268 & 95.0 \\
\ours & 4.5 & 1.5 & 54.5 & 0.623 & 0.801 & 22.153 & 0.273 & \textbf{95.6} \\
\ours & 4.5 & 2 & 54.5 & 0.618 & \textbf{0.802} & \textbf{22.163} & \textbf{0.275} & \textbf{95.6} \\
\midrule
\ours & 7 & 1 & 50.5 & 0.654 & 0.921 & 22.310 & 0.276 & 100.5 \\
\ours & 7 & 1.25 & 54.5 & \textbf{0.658} & \textbf{0.924} & 22.340 & 0.278 & \textbf{100.9} \\
\ours & 7 & 1.5 & 54.5 & 0.649 & \textbf{0.924} & \textbf{22.348} & 0.279 & 100.7 \\
\ours & 7 & 2 & 54.5 & 0.642 & 0.920 & 22.343 & \textbf{0.280} & 100.5 \\
\bottomrule
\end{tabular}
\end{table}

\subsection{Test-time samplers}
\label{app:samplers}

To test whether the depth-solver results of Tab.~\ref{tab:samplers_sd3} also hold on ImageNet and on PickScore and HPSv2.1, Tab.~\ref{tab:samplers_sit} reports the ImageNet counterpart of Tab.~\ref{tab:samplers_sd3}, with gFID and FD-DINOv2 of the SiT-XL/2 teacher and of \ours~for each time sampler and depth solver. Tab.~\ref{tab:samplers_sd3_pick_hps} reports PickScore and HPSv2.1 on the same images as Tab.~\ref{tab:samplers_sd3}.

\subsection{Adaptive loop counts}
\label{app:adaptive}

\begin{table}[tp]
\small
\centering
\caption{Choosing the number of loops per sampling step or per sample on the SD3 student, compared with fixed loop counts. Bold marks values above the teacher's, except in the fixed rows.}
\label{tab:adaptive}
\setlength{\tabcolsep}{1pt}
\begin{tabular}{lccccccc}
\toprule
Method & \shortstack{Mean\\loops} & TFLOPs & \shortstack{Gen-\\Eval$\uparrow$} & \shortstack{Image-\\Reward$\uparrow$} & \shortstack{Pick-\\Score$\uparrow$} & \shortstack{HPS-\\v2.1$\uparrow$} & \shortstack{Avg.\\(\%)$\uparrow$} \\
\midrule
Teacher & -- & 50.5 & 0.648 & 0.908 & 22.214 & 0.278 & 100.0 \\
\midrule
Fixed, $K{=}2$ & 2 & 29.4 & 0.662 & 0.922 & 22.243 & 0.275 & 100.7 \\
Fixed, $K{=}4$ & 4 & 50.5 & 0.654 & 0.921 & 22.310 & 0.276 & 100.5 \\
\midrule
\rowcolor{gray!15}\multicolumn{8}{l}{\textit{Looped-DiT exit gate~\cite{chng2026looped}}} \\
\hspace{3pt}$\tau{=}0.015$ & 2.01 & 29.5 & 0.553 & 0.529 & 21.476 & 0.246 & 82.2 \\
\midrule
\rowcolor{gray!15}\multicolumn{8}{l}{\textit{Exit gate with a finishing round}} \\
\hspace{3pt}$\tau{=}0.006$ & 2.64 & 35.7 & \textbf{0.661} & \textbf{0.926} & \textbf{22.272} & 0.275 & \textbf{100.8} \\
\hspace{3pt}$\tau{=}0.01$ & 2.13 & 30.8 & \textbf{0.652} & \textbf{0.930} & \textbf{22.227} & 0.273 & \textbf{100.4} \\
\hspace{3pt}$\tau{=}0.02$ & 1.51 & 24.2 & 0.637 & 0.854 & 22.089 & 0.269 & 97.2 \\
\hspace{3pt}$\tau{=}0.03$ & 1.25 & 21.5 & 0.620 & 0.788 & 21.964 & 0.265 & 94.2 \\
\midrule
\rowcolor{gray!15}\multicolumn{8}{l}{\textit{Loops per step, set before sampling}} \\
\hspace{3pt}Fit to teacher & 1.25 & 21.5 & 0.612 & 0.749 & 21.952 & 0.264 & 92.7 \\
\hspace{3pt}Fit to teacher & 1.5 & 24.1 & 0.629 & 0.811 & 22.039 & 0.268 & 95.5 \\
\hspace{3pt}Greedy & 1.25 & 21.5 & 0.612 & 0.762 & 21.926 & 0.264 & 93.0 \\
\hspace{3pt}Greedy & 1.5 & 24.1 & 0.632 & 0.818 & 22.057 & 0.267 & 95.8 \\
\hspace{3pt}Greedy & 1.75 & 26.8 & 0.633 & 0.833 & 22.117 & 0.269 & 96.4 \\
\midrule
\rowcolor{gray!15}\multicolumn{8}{l}{\textit{Loops per step, set by the previous step}} \\
\hspace{3pt}$\epsilon{=}0.3$ & 2.53 & 35.0 & \textbf{0.659} & \textbf{0.937} & \textbf{22.297} & 0.276 & \textbf{101.2} \\
\bottomrule
\end{tabular}
\end{table}

To test whether the loop count can change from step to step within one sample, Tab.~\ref{tab:adaptive} compares five ways to choose the number of loops.

\noindent\textbf{Looped-DiT exit gate.} This gate applies the per-sample exit gate of Looped-DiT~\cite{chng2026looped} to our student. A small gate network reads the image tokens after round $r$ together with the timestep embedding of the current step, through cross-attention with 4 learnable queries followed by an MLP, and predicts how much the error would drop if the loop continued. It is trained on the frozen student with the training objective of Looped-DiT, where the error is the relative squared error between the student's CFG-7 velocity after round $k$ and the teacher's CFG-7 velocity. At sampling time, within each denoising step, the loop stops after round $r = 1, 2, 3$ as soon as the predicted value is below the threshold $\tau$, and the depth then reaches only $\frac{r}{4}$.

\noindent\textbf{Exit gate with a finishing round.} This gate changes the Looped-DiT exit gate in two ways and is retrained. In the Looped-DiT gate, a sample that stops after round $r < 4$ gives its output at depth $\frac{r}{4}$. Here the stopped sample instead runs one more round of the shared blocks that takes it to depth 1. This round uses the depth coordinate $s = 1$ and the step of the interval from $\frac{r}{4}$ to 1, and the sample gives its output after it, which equals the forward on the non-uniform depth grid $\{\frac{1}{4}, \dots, \frac{r}{4}, 1\}$. Training already draws sorted random depth grids, so such grids are covered by training, and a sample that reaches round 4 is unchanged. The gate also decides right after the input blocks, before any shared round ($r = 0$), and stopping there runs a single round from the input-block state straight to depth 1, which is exactly the fixed $K = 1$ forward, so decisions are made at $r = 0, 1, 2, 3$. The gate keeps the architecture of the Looped-DiT gate, with a round embedding for $r = 0, \dots, 3$, and is trained on the same frozen student with the same data and number of steps. Its training errors are those of the finished outputs, $l_r = \frac{\lVert v^{\mathrm{fin}}_r - \vT \rVert^2}{\lVert \vT \rVert^2}$ for $r = 0, \dots, 4$, where $v^{\mathrm{fin}}_r$ is the CFG-7 velocity output after finishing from round $r$ ($r = 4$ is the full four-round output) and $\vT$ is the teacher's CFG-7 velocity. The target keeps the form of Looped-DiT, $y_r = \max\big(0, \max_{k = r+1, \dots, 4} \frac{l_r - l_k}{k - r}\big)$, and at sampling time the loop stops at the first $r$ whose predicted $y_r$ is below $\tau$. A sample that stops at $r$ runs $r + 1$ shared rounds in that step, namely $r$ rounds plus the finishing round, so the mean number of loops in Tab.~\ref{tab:adaptive} is the mean of these round counts over steps and images and includes the finishing round, and TFLOPs are computed from it as for the other rows.

\noindent\textbf{Loop counts set before sampling.} The sampler runs every denoising step $t$ with $K_t \in \{1, 2, 3, 4\}$ complete rounds on its own depth grid $s_k = \frac{k}{K_t}$, where the 28 values $K_t$ are given by a table and the mean of the table is the mean number of loops. The fit-to-teacher variant selects the table by dynamic programming, minimizing $\sum_t \Delta\sigma_t^2\, e(t, K_t)$ under a constraint on the mean number of loops. Here $e(t, K)$ is the squared error between the student's CFG-7 velocity with $K$ loops and the teacher's velocity, measured on the states of the $K=4$ sampling trajectory over 64 COCO captions. The greedy variant starts from a table of ones and repeatedly adds one round to one step, choosing the step for which the squared distance between the final latent and the teacher's final latent decreases most per added block, until the target mean is reached. It is searched on 32 COCO captions.

\noindent\textbf{Loop counts set by the previous step.} The sampler runs each step with $K_t$ complete rounds and records the relative change of the image tokens in the last round, $\delta_t = \frac{\lVert h_{K_t} - h_{K_t-1}\rVert^2}{\lVert h_{K_t-1}\rVert^2}$. This is converted into a change rate per unit depth, $\rho_t = \delta_t K_t^2$, and the next step uses $K_{t+1} = \min(4, \max(1, \lceil \sqrt{\rho_t / \epsilon}\,\rceil))$, with $K_1 = 4$. The value of $\epsilon$ is calibrated on 64 COCO captions. TFLOPs per image is the FLOPs of one function call multiplied by 28 steps, without the doubling for CFG, as in Tab.~\ref{tab:sd3}, and the compute of the gate network is not counted. The Fixed rows of Tab.~\ref{tab:adaptive} and Tab.~\ref{tab:sd3} come from separate generations.

\noindent\textbf{Results.} Setting the loops of the next step from the previous step with $\epsilon = 0.3$ uses a mean of 2.53 loops at about 69\% of the teacher's compute and is above the teacher on GenEval, ImageReward and PickScore, with GenEval 0.659 against 0.648. The Looped-DiT exit gate at a similar mean of 2.01 loops reaches GenEval 0.553. The loop count of \ours~can therefore also change from step to step within one sample.

\subsection{Continued training on real data after distillation}
\label{app:realdata}
\begin{figure*}[t]
\centering
\includegraphics[width=\textwidth]{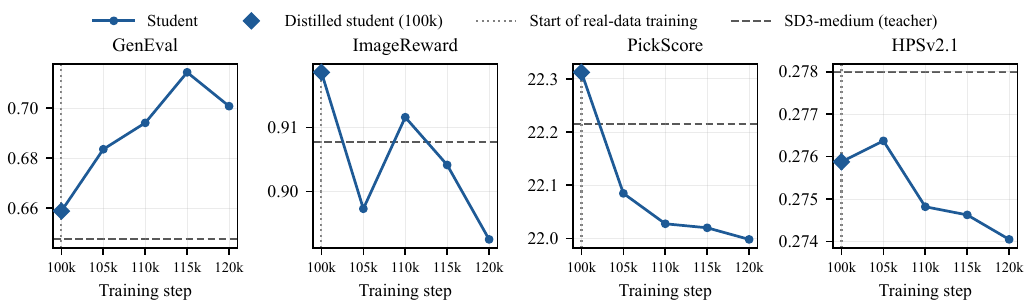}
\caption{GenEval, ImageReward, PickScore and HPSv2.1 of \ours~on SD3-medium during continued training on real data. The first point is the distilled student, and the dashed line marks the teacher.}
\label{fig:sd3_realdata}
\end{figure*}

To test whether the looped student can keep training with real data as the target after distillation, we continue training the distilled SD3 student, the 100k-step checkpoint of Tab.~\ref{tab:sd3}, from its full training state (weights, EMA, optimizer and learning-rate schedule) for 20k more steps. In this phase every loss term regresses the flow-matching velocity of real data, $\epsilon - x_0$, in place of the teacher's velocity, and neither teacher forward passes nor online rollouts are used. The data are the COCO-2017 train captions and latents of the distillation, with 144 rows per step, a constant learning rate of $1.732\times10^{-4}$ and an EMA decay of $0.9995$. Evaluation follows Tab.~\ref{tab:sd3} and uses EMA weights, 4 loops, 28 Euler steps and CFG 7 at 512 px, with the 553 GenEval prompts and 4 images each.

Fig.~\ref{fig:sd3_realdata} shows the results. Continuing on real data raises the GenEval overall score from 0.659 to 0.714 at 115k steps, above the teacher's 0.648, and raises Avg., the mean of the four metrics relative to the teacher as in Tab.~\ref{tab:sd3}, from 100.7 to 102.0. The largest gain among the GenEval categories is in position, from 0.273 to 0.378. ImageReward, PickScore and HPSv2.1 decrease slightly over the same steps.

\begin{table*}[t]
\footnotesize
\centering
\caption{gFID and FD-DINOv2 on ImageNet of the SiT-XL/2 teacher and of \ours~for each time sampler (rows) and each depth solver of the loop with $K$ loops (columns). The italic row gives the GFLOPs of one function evaluation (NFE), so the compute per sample is the number of NFEs times this value. Bold marks the lowest value among the \ours~columns in each row for each metric. Darker cells are better within each row and metric. The Heun sampler~\cite{karras2022elucidating} uses two function evaluations per step. DPM++ is DPM-Solver++~\cite{lu2022dpm,lu2025dpm}.}
\label{tab:samplers_sit}
\setlength{\tabcolsep}{1.3pt}
\begin{tabular}{lc@{\hspace{3pt}}cc@{\hspace{3pt}}cc@{\hspace{3pt}}cc@{\hspace{3pt}}cc@{\hspace{3pt}}cc@{\hspace{3pt}}cc@{\hspace{3pt}}cc@{\hspace{3pt}}cc}
\toprule
 & & & & \multicolumn{14}{c}{Depth solver of \ours} \\
\cmidrule(lr){5-18}
 & & & & \multicolumn{6}{c}{Euler} & \multicolumn{4}{c}{Heun} & \multicolumn{4}{c}{DPM++} \\
\cmidrule(lr){5-10}\cmidrule(lr){11-14}\cmidrule(lr){15-18}
 & & \multicolumn{2}{c}{Teacher} & \multicolumn{2}{c}{$K{=}2$} & \multicolumn{2}{c}{$K{=}4$} & \multicolumn{2}{c}{$K{=}8$} & \multicolumn{2}{c}{$K{=}2$} & \multicolumn{2}{c}{$K{=}4$} & \multicolumn{2}{c}{$K{=}2$} & \multicolumn{2}{c}{$K{=}4$} \\
\cmidrule(lr){3-4}\cmidrule(lr){5-6}\cmidrule(lr){7-8}\cmidrule(lr){9-10}\cmidrule(lr){11-12}\cmidrule(lr){13-14}\cmidrule(lr){15-16}\cmidrule(lr){17-18}
Sampler & Steps & gFID$\downarrow$ & FD$\downarrow$ & gFID$\downarrow$ & FD$\downarrow$ & gFID$\downarrow$ & FD$\downarrow$ & gFID$\downarrow$ & FD$\downarrow$ & gFID$\downarrow$ & FD$\downarrow$ & gFID$\downarrow$ & FD$\downarrow$ & gFID$\downarrow$ & FD$\downarrow$ & gFID$\downarrow$ & FD$\downarrow$ \\
\midrule
\textit{GFLOPs/NFE} & & \multicolumn{2}{c}{\textit{237.2}} & \multicolumn{2}{c}{\textit{135.6}} & \multicolumn{2}{c}{\textit{237.3}} & \multicolumn{2}{c}{\textit{440.6}} & \multicolumn{2}{c}{\textit{186.4}} & \multicolumn{2}{c}{\textit{389.8}} & \multicolumn{2}{c}{\textit{135.6}} & \multicolumn{2}{c}{\textit{237.3}} \\
\midrule
\multirow{3}{*}{Euler} & 10 & 9.897 & 230.846 & 11.657 & 347.916 & \cellcolor{ForestGreen!6}10.112 & \cellcolor{ForestGreen!6}278.935 & \cellcolor{ForestGreen!12}9.859 & \cellcolor{ForestGreen!19}262.406 & \cellcolor{ForestGreen!40}\textbf{9.299} & \cellcolor{ForestGreen!33}249.253 & \cellcolor{ForestGreen!26}9.484 & \cellcolor{ForestGreen!40}\textbf{247.111} & \cellcolor{ForestGreen!33}9.317 & \cellcolor{ForestGreen!12}276.981 & \cellcolor{ForestGreen!19}9.768 & \cellcolor{ForestGreen!26}260.166 \\
 & 25 & 5.443 & 163.411 & 7.412 & 278.992 & \cellcolor{ForestGreen!12}6.535 & \cellcolor{ForestGreen!12}215.101 & \cellcolor{ForestGreen!33}6.331 & \cellcolor{ForestGreen!19}204.240 & \cellcolor{ForestGreen!19}6.479 & \cellcolor{ForestGreen!33}190.673 & \cellcolor{ForestGreen!40}\textbf{6.330} & \cellcolor{ForestGreen!40}\textbf{188.463} & \cellcolor{ForestGreen!6}6.558 & \cellcolor{ForestGreen!6}215.624 & \cellcolor{ForestGreen!26}6.449 & \cellcolor{ForestGreen!26}199.244 \\
 & 50 & 4.774 & 147.844 & 6.860 & 262.188 & \cellcolor{ForestGreen!19}6.104 & \cellcolor{ForestGreen!12}198.762 & \cellcolor{ForestGreen!40}\textbf{5.972} & \cellcolor{ForestGreen!19}189.199 & \cellcolor{ForestGreen!12}6.162 & \cellcolor{ForestGreen!33}176.580 & \cellcolor{ForestGreen!33}6.059 & \cellcolor{ForestGreen!40}\textbf{174.109} & \cellcolor{ForestGreen!6}6.322 & \cellcolor{ForestGreen!6}201.427 & \cellcolor{ForestGreen!26}6.063 & \cellcolor{ForestGreen!26}182.832 \\
\cmidrule(l){2-18}
\multirow{2}{*}{Heun} & 10 & 5.375 & 168.340 & 7.138 & 280.702 & \cellcolor{ForestGreen!19}6.210 & \cellcolor{ForestGreen!12}217.422 & \cellcolor{ForestGreen!33}6.041 & \cellcolor{ForestGreen!19}207.450 & \cellcolor{ForestGreen!12}6.239 & \cellcolor{ForestGreen!33}194.430 & \cellcolor{ForestGreen!40}\textbf{6.002} & \cellcolor{ForestGreen!40}\textbf{191.111} & \cellcolor{ForestGreen!6}6.261 & \cellcolor{ForestGreen!6}217.836 & \cellcolor{ForestGreen!26}6.131 & \cellcolor{ForestGreen!26}201.177 \\
 & 25 & 4.603 & 141.674 & 6.682 & 253.590 & \cellcolor{ForestGreen!33}5.920 & \cellcolor{ForestGreen!12}192.588 & \cellcolor{ForestGreen!40}\textbf{5.817} & \cellcolor{ForestGreen!19}183.843 & \cellcolor{ForestGreen!12}6.118 & \cellcolor{ForestGreen!33}172.179 & \cellcolor{ForestGreen!19}5.970 & \cellcolor{ForestGreen!40}\textbf{170.249} & \cellcolor{ForestGreen!6}6.226 & \cellcolor{ForestGreen!6}195.622 & \cellcolor{ForestGreen!26}5.953 & \cellcolor{ForestGreen!26}177.191 \\
\cmidrule(l){2-18}
\multirow{2}{*}{DPM++} & 10 & 6.061 & 184.301 & 8.087 & 304.373 & \cellcolor{ForestGreen!6}7.116 & \cellcolor{ForestGreen!12}238.342 & \cellcolor{ForestGreen!33}7.006 & \cellcolor{ForestGreen!19}225.827 & \cellcolor{ForestGreen!26}7.039 & \cellcolor{ForestGreen!33}216.670 & \cellcolor{ForestGreen!19}7.044 & \cellcolor{ForestGreen!40}\textbf{214.447} & \cellcolor{ForestGreen!12}7.046 & \cellcolor{ForestGreen!6}239.535 & \cellcolor{ForestGreen!40}\textbf{6.995} & \cellcolor{ForestGreen!26}221.954 \\
 & 25 & 4.748 & 148.635 & 6.767 & 264.343 & \cellcolor{ForestGreen!26}5.957 & \cellcolor{ForestGreen!12}202.672 & \cellcolor{ForestGreen!40}\textbf{5.890} & \cellcolor{ForestGreen!19}194.336 & \cellcolor{ForestGreen!12}6.095 & \cellcolor{ForestGreen!33}180.449 & \cellcolor{ForestGreen!19}5.972 & \cellcolor{ForestGreen!40}\textbf{179.120} & \cellcolor{ForestGreen!6}6.190 & \cellcolor{ForestGreen!6}203.756 & \cellcolor{ForestGreen!33}5.950 & \cellcolor{ForestGreen!26}186.434 \\
\bottomrule
\end{tabular}
\end{table*}

\begin{table*}[t]
\footnotesize
\centering
\caption{PickScore and HPSv2.1 of the SD3-medium teacher and of \ours~for each time sampler (rows) and each depth solver of the loop with $K$ loops (columns). The italic row gives the GFLOPs of one function evaluation (NFE). Bold marks the highest value among the \ours~columns in each row for each metric. Darker cells are better within each row and metric.}
\label{tab:samplers_sd3_pick_hps}
\setlength{\tabcolsep}{1.3pt}
\begin{tabular}{lc@{\hspace{3pt}}cc@{\hspace{3pt}}cc@{\hspace{3pt}}cc@{\hspace{3pt}}cc@{\hspace{3pt}}cc@{\hspace{3pt}}cc@{\hspace{3pt}}cc@{\hspace{3pt}}cc}
\toprule
 & & & & \multicolumn{14}{c}{Depth solver of \ours} \\
\cmidrule(lr){5-18}
 & & & & \multicolumn{6}{c}{Euler} & \multicolumn{4}{c}{Heun} & \multicolumn{4}{c}{DPM++} \\
\cmidrule(lr){5-10}\cmidrule(lr){11-14}\cmidrule(lr){15-18}
 & & \multicolumn{2}{c}{Teacher} & \multicolumn{2}{c}{$K{=}2$} & \multicolumn{2}{c}{$K{=}4$} & \multicolumn{2}{c}{$K{=}8$} & \multicolumn{2}{c}{$K{=}2$} & \multicolumn{2}{c}{$K{=}4$} & \multicolumn{2}{c}{$K{=}2$} & \multicolumn{2}{c}{$K{=}4$} \\
\cmidrule(lr){3-4}\cmidrule(lr){5-6}\cmidrule(lr){7-8}\cmidrule(lr){9-10}\cmidrule(lr){11-12}\cmidrule(lr){13-14}\cmidrule(lr){15-16}\cmidrule(lr){17-18}
Sampler & Steps & \shortstack{Pick-\\Score$\uparrow$} & \shortstack{HPS-\\v2.1$\uparrow$} & \shortstack{Pick-\\Score$\uparrow$} & \shortstack{HPS-\\v2.1$\uparrow$} & \shortstack{Pick-\\Score$\uparrow$} & \shortstack{HPS-\\v2.1$\uparrow$} & \shortstack{Pick-\\Score$\uparrow$} & \shortstack{HPS-\\v2.1$\uparrow$} & \shortstack{Pick-\\Score$\uparrow$} & \shortstack{HPS-\\v2.1$\uparrow$} & \shortstack{Pick-\\Score$\uparrow$} & \shortstack{HPS-\\v2.1$\uparrow$} & \shortstack{Pick-\\Score$\uparrow$} & \shortstack{HPS-\\v2.1$\uparrow$} & \shortstack{Pick-\\Score$\uparrow$} & \shortstack{HPS-\\v2.1$\uparrow$} \\
\midrule
\textit{GFLOPs/NFE} & & \multicolumn{2}{c}{\textit{1802.6}} & \multicolumn{2}{c}{\textit{1049.8}} & \multicolumn{2}{c}{\textit{1802.7}} & \multicolumn{2}{c}{\textit{3308.4}} & \multicolumn{2}{c}{\textit{1426.2}} & \multicolumn{2}{c}{\textit{2931.9}} & \multicolumn{2}{c}{\textit{1049.8}} & \multicolumn{2}{c}{\textit{1802.7}} \\
\midrule
\multirow{3}{*}{Euler} & 10 & 21.495 & 0.252 & \cellcolor{ForestGreen!19}21.804 & \cellcolor{ForestGreen!19}0.257 & \cellcolor{ForestGreen!40}\textbf{21.868} & \cellcolor{ForestGreen!26}0.258 & \cellcolor{ForestGreen!26}21.813 & \cellcolor{ForestGreen!12}0.256 & 21.685 & \cellcolor{ForestGreen!33}0.260 & \cellcolor{ForestGreen!6}21.721 & \cellcolor{ForestGreen!12}0.256 & \cellcolor{ForestGreen!33}21.830 & \cellcolor{ForestGreen!40}\textbf{0.261} & \cellcolor{ForestGreen!12}21.740 & \cellcolor{ForestGreen!12}0.256 \\
 & 25 & 22.177 & 0.277 & \cellcolor{ForestGreen!12}22.220 & \cellcolor{ForestGreen!19}0.275 & \cellcolor{ForestGreen!40}\textbf{22.317} & \cellcolor{ForestGreen!26}0.277 & \cellcolor{ForestGreen!33}22.271 & \cellcolor{ForestGreen!19}0.275 & 22.135 & \cellcolor{ForestGreen!40}\textbf{0.278} & \cellcolor{ForestGreen!19}22.222 & \cellcolor{ForestGreen!19}0.275 & \cellcolor{ForestGreen!26}22.252 & \cellcolor{ForestGreen!40}\textbf{0.278} & \cellcolor{ForestGreen!6}22.206 & \cellcolor{ForestGreen!19}0.275 \\
 & 50 & 22.369 & 0.283 & \cellcolor{ForestGreen!12}22.295 & 0.277 & \cellcolor{ForestGreen!40}\textbf{22.372} & \cellcolor{ForestGreen!26}0.279 & \cellcolor{ForestGreen!33}22.332 & \cellcolor{ForestGreen!12}0.278 & 22.193 & \cellcolor{ForestGreen!40}\textbf{0.281} & \cellcolor{ForestGreen!19}22.296 & \cellcolor{ForestGreen!12}0.278 & \cellcolor{ForestGreen!26}22.299 & \cellcolor{ForestGreen!33}0.280 & \cellcolor{ForestGreen!6}22.284 & \cellcolor{ForestGreen!26}0.279 \\
\cmidrule(l){2-18}
\multirow{2}{*}{Heun} & 10 & 21.738 & 0.262 & \cellcolor{ForestGreen!26}21.952 & \cellcolor{ForestGreen!26}0.265 & \cellcolor{ForestGreen!40}\textbf{21.987} & \cellcolor{ForestGreen!26}0.265 & \cellcolor{ForestGreen!19}21.936 & \cellcolor{ForestGreen!12}0.264 & 21.828 & \cellcolor{ForestGreen!40}\textbf{0.268} & \cellcolor{ForestGreen!6}21.877 & 0.263 & \cellcolor{ForestGreen!33}21.961 & \cellcolor{ForestGreen!40}\textbf{0.268} & \cellcolor{ForestGreen!12}21.889 & \cellcolor{ForestGreen!12}0.264 \\
 & 25 & 22.236 & 0.279 & \cellcolor{ForestGreen!19}22.224 & \cellcolor{ForestGreen!19}0.275 & \cellcolor{ForestGreen!40}\textbf{22.296} & \cellcolor{ForestGreen!26}0.276 & \cellcolor{ForestGreen!33}22.245 & \cellcolor{ForestGreen!19}0.275 & 22.091 & \cellcolor{ForestGreen!40}\textbf{0.278} & \cellcolor{ForestGreen!12}22.217 & \cellcolor{ForestGreen!19}0.275 & \cellcolor{ForestGreen!26}22.242 & \cellcolor{ForestGreen!40}\textbf{0.278} & \cellcolor{ForestGreen!6}22.180 & \cellcolor{ForestGreen!19}0.275 \\
\cmidrule(l){2-18}
\multirow{2}{*}{DPM++} & 10 & 20.850 & 0.239 & \cellcolor{ForestGreen!26}21.520 & \cellcolor{ForestGreen!33}0.257 & \cellcolor{ForestGreen!40}\textbf{21.603} & \cellcolor{ForestGreen!40}\textbf{0.259} & \cellcolor{ForestGreen!33}21.530 & \cellcolor{ForestGreen!19}0.256 & 21.264 & \cellcolor{ForestGreen!12}0.254 & \cellcolor{ForestGreen!12}21.408 & 0.253 & \cellcolor{ForestGreen!19}21.455 & \cellcolor{ForestGreen!33}0.257 & \cellcolor{ForestGreen!6}21.381 & \cellcolor{ForestGreen!12}0.254 \\
 & 25 & 21.936 & 0.268 & \cellcolor{ForestGreen!19}22.107 & \cellcolor{ForestGreen!19}0.272 & \cellcolor{ForestGreen!40}\textbf{22.182} & \cellcolor{ForestGreen!26}0.274 & \cellcolor{ForestGreen!26}22.122 & \cellcolor{ForestGreen!19}0.272 & 22.003 & \cellcolor{ForestGreen!40}\textbf{0.275} & \cellcolor{ForestGreen!12}22.063 & \cellcolor{ForestGreen!6}0.271 & \cellcolor{ForestGreen!33}22.136 & \cellcolor{ForestGreen!40}\textbf{0.275} & \cellcolor{ForestGreen!6}22.057 & \cellcolor{ForestGreen!6}0.271 \\
\bottomrule
\end{tabular}
\end{table*}

\subsection{Results on Z-Image}
\label{app:zimage}
\begin{table*}[t]
\small
\centering
\caption{Text-to-image generation with Z-Image as the teacher. Best student result of each part in bold.}
\label{tab:zimage}
\setlength{\tabcolsep}{2pt}
\begin{tabular}{lcccccccccccc}
\toprule
 & & \multicolumn{7}{c}{GenEval~\cite{ghosh2023geneval}$\uparrow$} & & & & \\
\cmidrule(lr){3-9}
Model & Rounds & \shortstack{Single\\obj.} & \shortstack{Two\\obj.} & Counting & Colors & Position & \shortstack{Color\\attr.} & Overall & \shortstack{Image-\\Reward\\\cite{xu2023imagereward}$\uparrow$} & \shortstack{Pick-\\Score\\\cite{kirstain2023pick}$\uparrow$} & \shortstack{HPS-\\v2.1\\\cite{wu2023human}$\uparrow$} & \shortstack{Avg.\\(\%)}$\uparrow$ \\
\midrule
Z-Image~\cite{zimage2025} (teacher) & -- & 0.994 & 0.861 & 0.641 & 0.854 & 0.347 & 0.535 & 0.705 & 0.947 & 22.742 & 0.286 & 100.0 \\
\midrule
\rowcolor{gray!15}\multicolumn{13}{l}{\textit{Distill}} \\
\quad \ours~(Ours) & 1 & 0.991 & 0.765 & 0.466 & 0.830 & 0.400 & 0.557 & 0.668 & 0.847 & 22.156 & 0.275 & 94.4 \\
\quad \ours~(Ours) & 2 & \textbf{0.997} & 0.833 & 0.553 & 0.854 & \textbf{0.412} & 0.613 & 0.710 & 0.991 & 22.468 & 0.283 & 100.7 \\
\quad \ours~(Ours) & 3 & 0.994 & 0.876 & 0.566 & 0.867 & 0.380 & 0.652 & 0.723 & 1.023 & 22.547 & \textbf{0.285} & 102.3 \\
\quad \ours~(Ours) & 4 & 0.994 & 0.874 & 0.591 & 0.872 & 0.375 & \textbf{0.660} & 0.728 & \textbf{1.026} & \textbf{22.556} & \textbf{0.285} & 102.6 \\
\quad \ours~(Ours) & 8 & \textbf{0.997} & \textbf{0.891} & 0.591 & \textbf{0.875} & 0.410 & 0.655 & \textbf{0.736} & 1.025 & 22.548 & \textbf{0.285} & \textbf{102.9} \\
\quad \ours~(Ours) & 16 & \textbf{0.997} & 0.884 & \textbf{0.619} & 0.867 & 0.388 & 0.627 & 0.730 & 1.012 & 22.512 & \textbf{0.285} & 102.3 \\
\midrule
\rowcolor{gray!15}\multicolumn{13}{l}{+~\textit{New data}} \\
\quad \ours~(Ours) & 1 & 0.994 & 0.758 & 0.469 & 0.838 & 0.453 & 0.565 & 0.679 & 0.840 & 22.201 & 0.276 & 94.8 \\
\quad \ours~(Ours) & 2 & \textbf{0.997} & 0.854 & 0.562 & \textbf{0.851} & \textbf{0.465} & \textbf{0.640} & \textbf{0.728} & 0.979 & 22.479 & 0.283 & 101.1 \\
\quad \ours~(Ours) & 3 & \textbf{0.997} & 0.848 & 0.569 & \textbf{0.851} & 0.453 & \textbf{0.640} & 0.726 & 1.012 & 22.561 & \textbf{0.285} & 102.2 \\
\quad \ours~(Ours) & 4 & 0.994 & \textbf{0.869} & 0.562 & 0.843 & 0.443 & 0.635 & 0.724 & \textbf{1.022} & \textbf{22.575} & \textbf{0.285} & \textbf{102.4} \\
\quad \ours~(Ours) & 5 & 0.991 & \textbf{0.869} & 0.591 & 0.824 & 0.438 & 0.627 & 0.723 & 1.019 & 22.569 & \textbf{0.285} & 102.3 \\
\quad \ours~(Ours) & 6 & 0.991 & \textbf{0.869} & \textbf{0.597} & 0.832 & 0.440 & 0.618 & 0.724 & 1.009 & 22.557 & \textbf{0.285} & 102.0 \\
\midrule
\rowcolor{gray!15}\multicolumn{13}{l}{+~\textit{Student rollout}} \\
\quad \ours~(Ours) & 1 & 0.997 & 0.770 & 0.450 & 0.832 & 0.372 & 0.562 & 0.664 & 0.884 & 22.325 & 0.275 & 95.5 \\
\quad \ours~(Ours) & 2 & \textbf{1.000} & 0.881 & 0.597 & \textbf{0.862} & \textbf{0.407} & \textbf{0.662} & 0.735 & 1.046 & 22.645 & 0.284 & 103.4 \\
\quad \ours~(Ours) & 3 & \textbf{1.000} & 0.881 & 0.591 & 0.843 & 0.400 & 0.647 & 0.727 & 1.063 & \textbf{22.716} & \textbf{0.285} & 103.7 \\
\quad \ours~(Ours) & 4 & \textbf{1.000} & 0.899 & 0.628 & 0.838 & \textbf{0.407} & 0.647 & \textbf{0.737} & 1.065 & 22.705 & \textbf{0.285} & \textbf{104.1} \\
\quad \ours~(Ours) & 5 & 0.997 & 0.914 & 0.637 & 0.835 & 0.393 & 0.627 & 0.734 & \textbf{1.070} & 22.705 & \textbf{0.285} & \textbf{104.1} \\
\quad \ours~(Ours) & 6 & 0.997 & \textbf{0.917} & \textbf{0.653} & 0.824 & 0.395 & 0.610 & 0.733 & 1.062 & 22.700 & \textbf{0.285} & 103.9 \\
\bottomrule
\end{tabular}
\end{table*}

To test whether \ours~carries over to a larger teacher, we use Z-Image~\cite{zimage2025} as the teacher, a 6.15B single-stream diffusion transformer with 30 blocks. The \ours~student has 3 input, 6 looped and 3 output blocks and 2.90B trainable parameters, and is distilled for 20k steps on COCO 2017 train at 512 px with the hyperparameters of Tab.~\ref{tab:hparams}. From the 20k-step checkpoint with its full training state, we continue training it for 5k more steps in two separate ways. In \textit{+ new data}, the target stays the teacher's velocity, but the COCO data are replaced by 181,107 real images drawn from PD12M~\cite{meyer2024public}, Open Images~\cite{kuznetsova2020open} and TextOCR~\cite{singh2021textocr}, covering paintings, photographs, people, scenes with countable objects, drawings and prints, and text in images, with captions written by a vision-language model. In \textit{+ student rollout}, the data stay COCO, but the online rollout states, which the distillation recipe takes from the sampler that combines the teacher and the current student, come from the student's own sampling trajectory. The student samples from pure noise with four loops, the evaluation sampler with 28 steps and the evaluation CFG, and 8 states per trajectory are drawn uniformly from steps 1 to 27. Their targets are the same as for data states.

Evaluation follows Tab.~\ref{tab:sd3}: the 553 GenEval prompts with 4 images each at 512 px and 28 sampling steps, with ImageReward, PickScore and HPSv2.1 on the same images, and Avg. is the mean of the four metrics, each divided by the teacher's value. Tab.~\ref{tab:zimage} reports the results. The distilled student already exceeds the teacher at two loops, with an Avg. of 100.7 against 100.0, and the \textit{+ student rollout} student reaches an Avg. of 104.1 at four loops. Fig.~\ref{fig:qual_zimage_rollout} and Fig.~\ref{fig:qual_zimage_newdata} show samples of the two continued students.

\section{Additional qualitative results}
\label{app:qual}

\begin{figure*}[p]
\centering
\includegraphics[width=\textwidth]{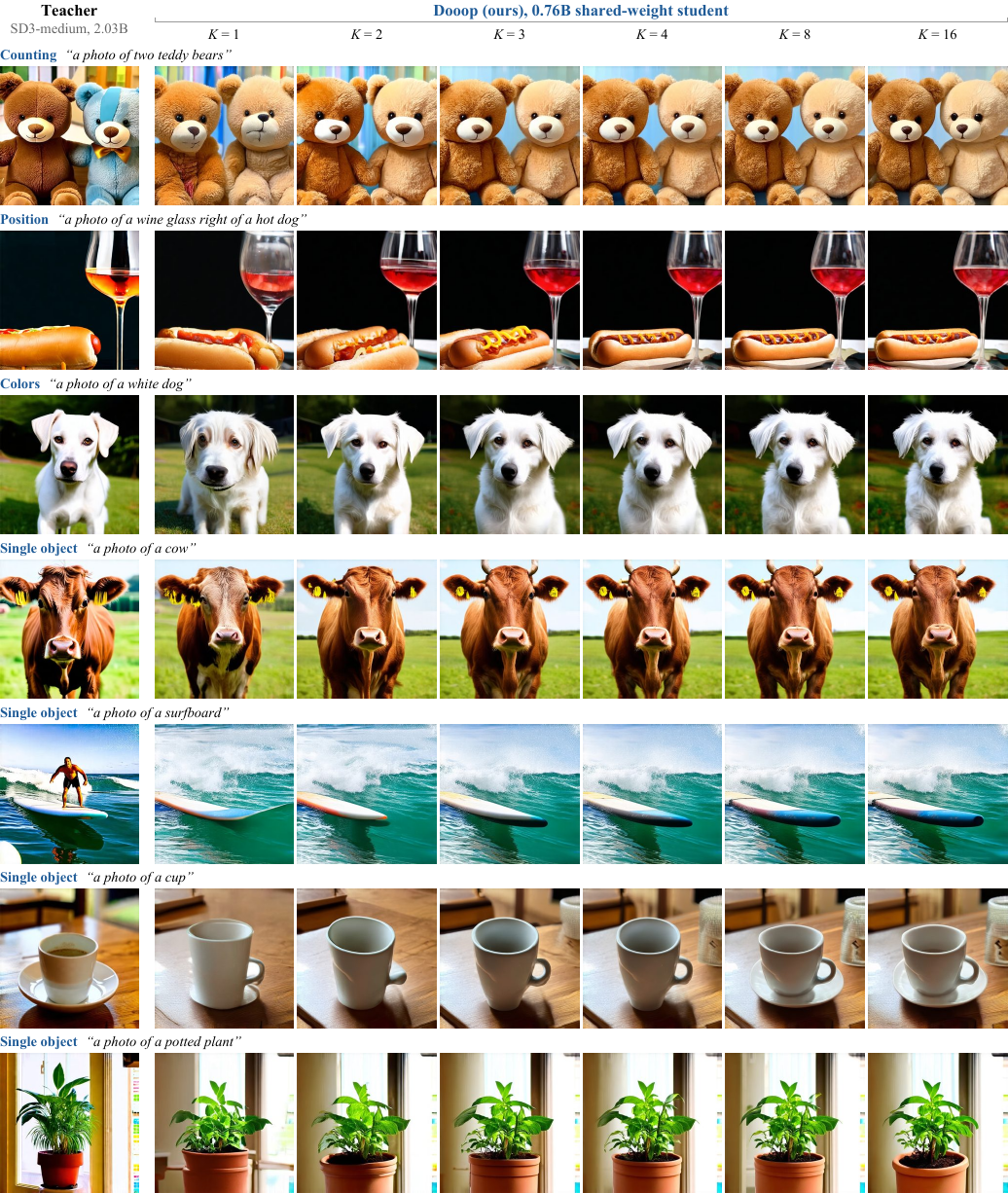}
\caption{More samples of the SD3-medium teacher and of \ours~with $K = 1, 2, 3, 4, 8, 16$ loops on GenEval prompts. All images in a row start from the same initial noise.}
\label{fig:qual_sd3_appA}
\end{figure*}

\begin{figure*}[p]
\centering
\includegraphics[width=\textwidth]{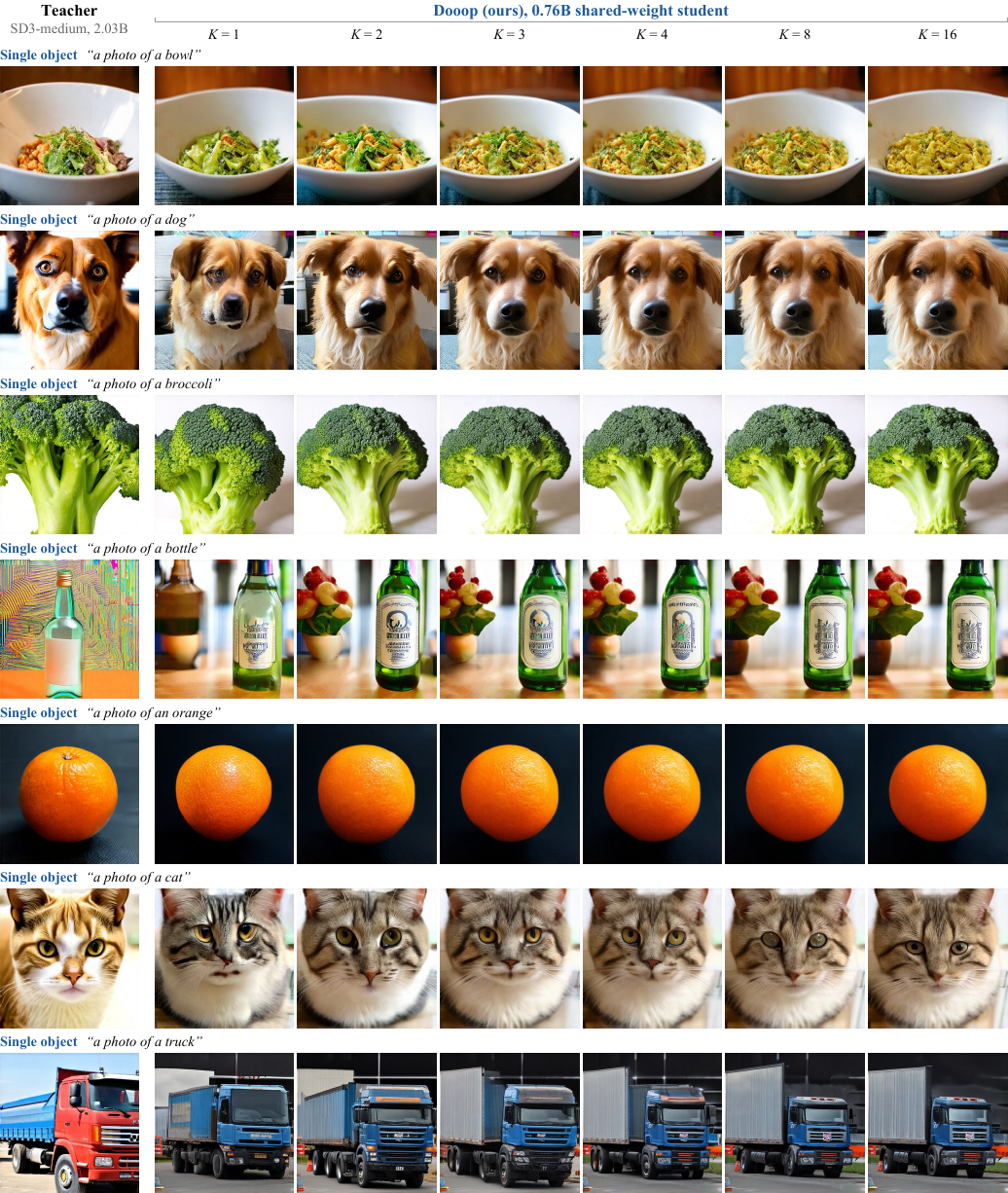}
\caption{More samples of the SD3-medium teacher and of \ours~with $K = 1, 2, 3, 4, 8, 16$ loops on GenEval single-object prompts. All images in a row start from the same initial noise.}
\label{fig:qual_sd3_appB}
\end{figure*}

\begin{figure*}[p]
\centering
\includegraphics[width=\textwidth,height=0.92\textheight,keepaspectratio]{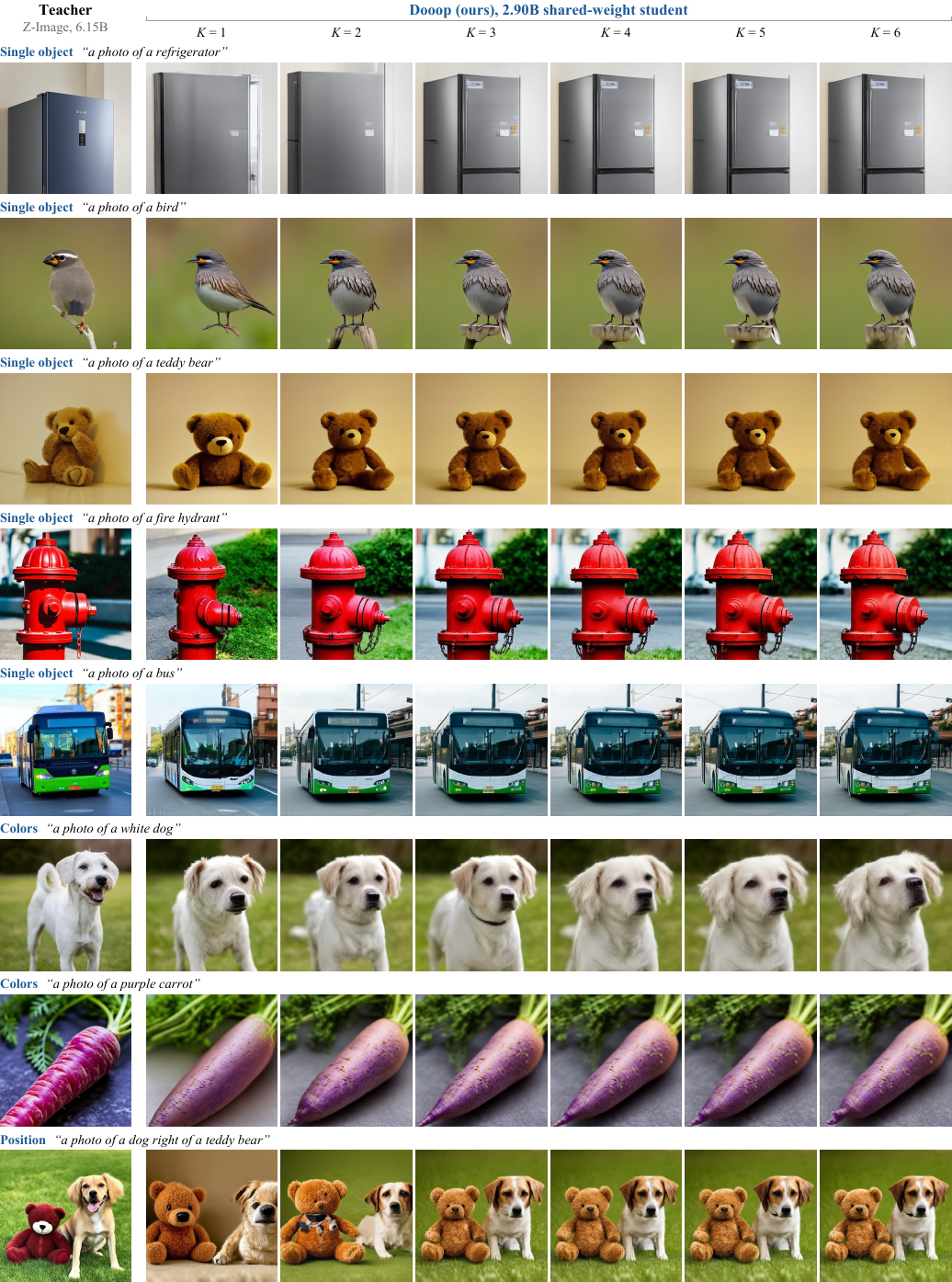}
\caption{More samples of the Z-Image teacher and of the \ours~student continued with student rollout, with $K = 1$ to $6$ loops on GenEval prompts. All images in a row start from the same initial noise.}
\label{fig:qual_zimage_rollout}
\end{figure*}

\begin{figure*}[p]
\centering
\includegraphics[width=\textwidth,height=0.92\textheight,keepaspectratio]{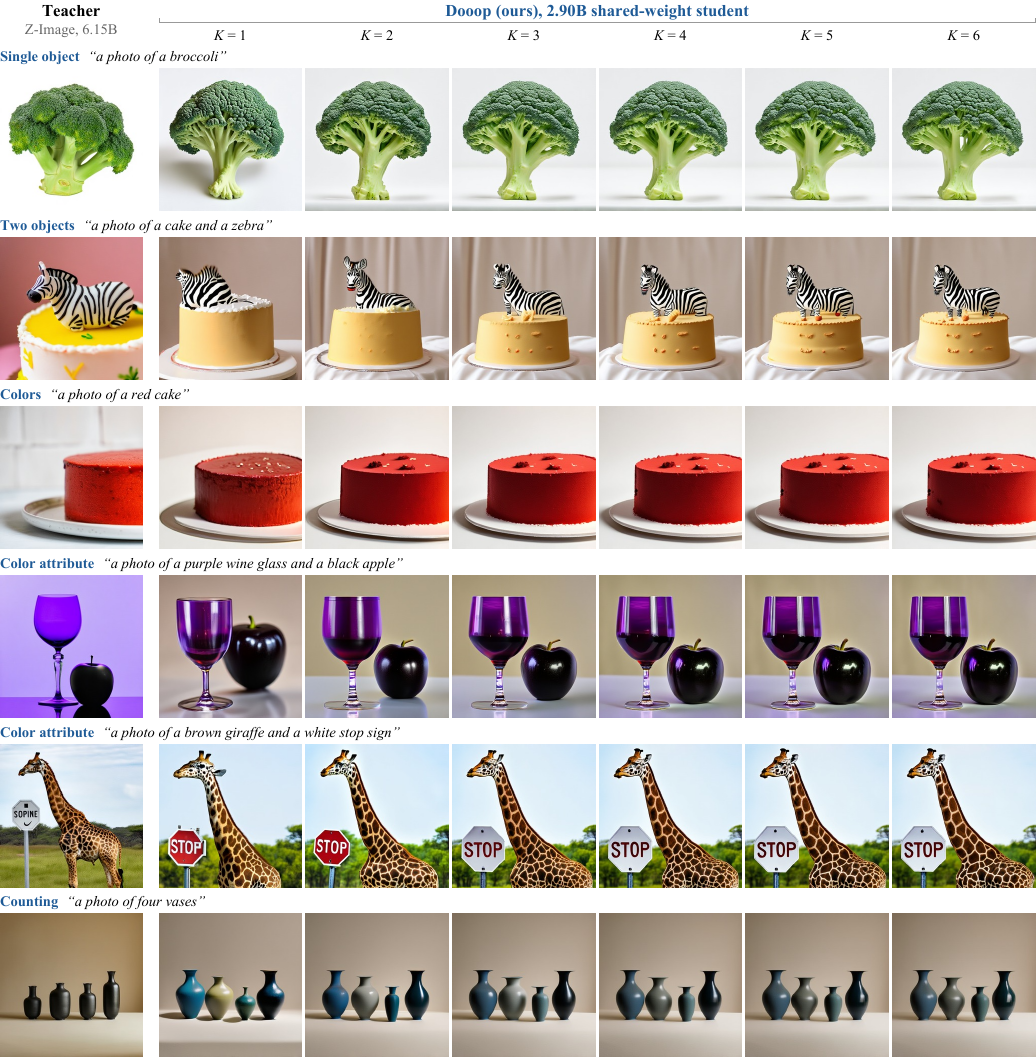}
\caption{More samples of the Z-Image teacher and of the \ours~student continued with new data, with $K = 1$ to $6$ loops on GenEval prompts. All images in a row start from the same initial noise.}
\label{fig:qual_zimage_newdata}
\end{figure*}

Fig.~\ref{fig:qual_sd3_appA} and Fig.~\ref{fig:qual_sd3_appB} show more samples of the SD3-medium teacher and of \ours~on GenEval prompts. Fig.~\ref{fig:teaser_zimage} shows samples of the Z-Image student on long text prompts, and Fig.~\ref{fig:qual_zimage_rollout} and Fig.~\ref{fig:qual_zimage_newdata} show samples of the Z-Image teacher and of the two continued Z-Image students.



\section{Limitations and future work}
\label{app:future}

Distillation into the looped architecture still needs a long training run, since the SD3-medium student is distilled for 100k steps. A natural next step is to shorten training, for example by combining loop distillation with a distribution matching objective~\cite{yin2024one,yin2024improved}. Our few-step results obtain a four-step looped student by distilling the four-step SD3-medium model of CDM~\cite{liu2026continuous}, so they rely on a separately trained few-step teacher. Applying step distillation directly to a looped student distilled from the multi-step teacher would let loop distillation and step distillation compose, so a few-step looped model would no longer need a separate few-step teacher. Another limitation is that we have not yet applied \ours~to text-to-video diffusion models. A natural next step is to apply it to models such as Wan2.2-TI2V-5B~\cite{wan2025wan}, since video generation is where the inference cost is highest.

\end{document}